\pdfoutput=1

\documentclass[11pt]{article}

\usepackage[final]{acl}

\usepackage{times}
\usepackage{latexsym}
\usepackage{tabularx}
\usepackage{float}
\usepackage{lipsum}
\usepackage{latexsym}
\usepackage{graphicx}
\usepackage{amsmath}
\usepackage{amssymb}
\usepackage{bm}
\usepackage{subcaption}
\usepackage{multirow}
\usepackage{enumitem}
\setlist{topsep=1pt,itemsep=1pt,partopsep=1pt, parsep=1pt}
\usepackage{makecell}

\usepackage[T1]{fontenc}

\usepackage[utf8]{inputenc}

\usepackage{microtype}

\usepackage{inconsolata}

\usepackage{graphicx}

\usepackage{comment}

\usepackage{booktabs}
\usepackage{array}
\usepackage{colortbl} 
\usepackage{xcolor} 
\usepackage{adjustbox} 

\newcommand{\methodname}[1]{\texttt{ITLC}}

\title{Language Surgery in Multilingual Large Language Models}

\author {
    Joanito Agili Lopo$^{*1,2}$, Muhammad Ravi Shulthan Habibi$^{*1,3}$, Tack Hwa Wong$^{*1}$, \\
    \textbf{Muhammad Ilham Ghozali$^{1,3}$, Fajri Koto$^{1,4}$, Genta Indra Winata}$^{1,5}$, \\
    \textbf{Peerat Limkonchotiwat$^{1,6}$, Alham Fikri Aji$^{1,4}$, Samuel Cahyawijaya\thanks{Equal contributions.  See Appendix~\ref{app:author_contribution} for further details.}$^{1,7}$} \\
    $^1$SEACrowd$\quad$$^2$Kreasof AI$\quad$$^3$Universitas Indonesia\\
    $^4$MBZUAI$\quad$$^5$Capital One$\quad$ $^6$AI Singapore$\quad$$^7$Cohere \\
    \texttt{\{amalopo99,muhammadravi251001,tackhwawong00\}@gmail.com} \\ 
    \texttt{muhammad.ilham.gozali@gmail.com,samuelcahyawijaya@cohere.com} \\ 
    Code: \url{https://github.com/SEACrowd/itlc} \\
    \\
}

\begin{document}
\maketitle

\begin{abstract}
Large Language Models (LLMs) have demonstrated remarkable generalization capabilities across tasks and languages, revolutionizing natural language processing. This paper investigates the naturally emerging representation alignment in LLMs, particularly in the middle layers, and its implications for disentangling language-specific and language-agnostic information. We empirically confirm the existence of this alignment, analyze its behavior in comparison to explicitly designed alignment models, and demonstrate its potential for language-specific manipulation without semantic degradation. Building on these findings, we propose Inference-Time Language Control (\methodname{}), a novel method that leverages latent injection to enable precise cross-lingual language control and mitigate language confusion in LLMs. Our experiments highlight \methodname{}'s strong cross-lingual control capabilities while preserving semantic integrity in target languages. Furthermore, we demonstrate its effectiveness in alleviating the cross-lingual language confusion problem, which persists even in current large-scale LLMs, leading to inconsistent language generation. This work advances our understanding of representation alignment in LLMs and introduces a practical solution for enhancing their monolingual and cross-lingual performance.

\end{abstract}

\section{Introduction}

Large Language Models (LLMs) have revolutionized natural language processing, demonstrating remarkable generalization capabilities across diverse tasks and languages \citep{brown2020language,le2023bloom,anil2023palm,google2025gemma3,cohere2025commanda,singh2025globalmmluunderstandingaddressing}. Their ability to adapt to new tasks in few-shot and even zero-shot settings highlights their efficiency and versatility~\citep{bang-etal-2023-multitask,susanto2025seahelmsoutheastasianholistic}. Prior works have identified a naturally emerging representation alignment across layers in LLMs, particularly in the middle layers of LLMs~\citep{chang-etal-2022-geometry,zhao2024how}. This emerging alignment in LLMs is the key factor in their ability to handle multiple languages~\cite{cahyawijaya2024llm4everyone,tang-etal-2024-language,wilie2025interlingua},
which is pivotal for their cross-lingual capabilities. However, several questions remain open, such as whether this emerging alignment behaves similarly to alignment in models trained with enforced alignment objectives~\citep{reimers-gurevych-2020-making,yang2019improving,feng2022languageagnosticbertsentenceembedding,limkonchotiwat-etal-2022-cl,limkonchotiwat-etal-2024-mccrolin}, how this alignment can be utilized to further enhance LLMs, etc.

\begin{figure}[!t]
    \centering
    \includegraphics[width=\linewidth]{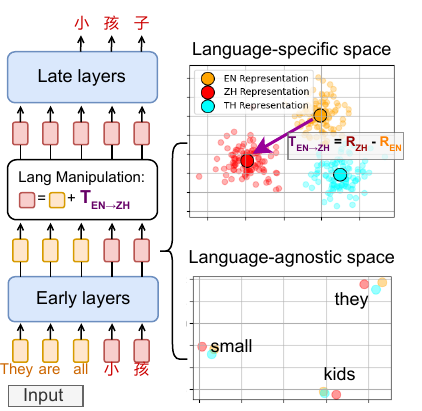}
    \caption{We inspect the alignment in the middle layer representation of LLMs, allowing us to disentangle the language-specific and language-agnostic information. By exploiting this behavior, we are able to achieve Inference-Time Language Control (\methodname{}), alleviating the language confusion problem in LLMs.}
    \label{fig:overview}
    \vspace{-3pt}
\end{figure}

In this work, we investigate the phenomenon of representation alignment in LLMs, focusing on its occurrence, distinction, and potential applications. We aim to confirm the presence of representation alignment and contrast it with alignment in LLMs with strictly designed alignment, such as multilingual SentenceBERT~\cite{reimers-2019-sentence-bert} or LaBSE~\cite{feng2022languageagnosticbertsentenceembedding}. Our findings highlight that, unlike LLMs with strictly designed alignment, the naturally emerging alignment in recent LLMs demonstrates a much stronger retention of language-specific information with much smaller performance drop in the aligned representation compared to the unaligned layers which we conjecture to be the minimum required language-specific information required to perform do decoding in the correct language.

To this end, we exploit the bottleneck of language-specific information in the aligned representation and develop a simple test-time intervention method to control the decoding language, namely inference-time language control (\methodname{}). Specifically, we extracted a low-rank language vector from the aligned representations using linear discriminant analysis~\cite{balakrishnama1998lda,tharwat2017lda}, aggregated them per language to create language vectors, and perform a simple vector translation to control the decoding language as shown in Figure~\ref{fig:overview}~\footnote{Note that, during the inference step, we only need to perform a single vector addition operation to control the language as everything else can be precomputed.}. We show the effectiveness of \methodname{} in mitigating the language confusion problem~\cite{marchisio-etal-2024-understanding}. Furthermore, we conduct an extensive evaluation to test that, unlike other approaches, \methodname{} can control the language with minimal loss of semantic.

Our contribution in this work is fourfold:
\begin{itemize}
\item We confirm the presence of representation alignment in LLMs, providing empirical evidence of this phenomenon (\S\ref{sec:understanding-result}).
\item We contrast natural alignment with strictly designed alignment, highlighting their comparable impact on cross-lingual generalization while emphasizing their differences in alignment locations and the extent of language-specific information retention (\S\ref{sec:understanding-result}).
\item We investigate a method to extract language-specific information from aligned representations, showcasing the potential for language-specific manipulation while preserving the semantic integrity of the generation (\S\ref{sec:method}).
\item We introduce \methodname{}, a novel method that enables cross-lingual language control and mitigates language confusion problems that retain semantic integrity in target languages (\S\ref{sec:implication}).
\end{itemize}

\begin{figure*}[!t]
    \centering
    \includegraphics[clip,width=\textwidth]{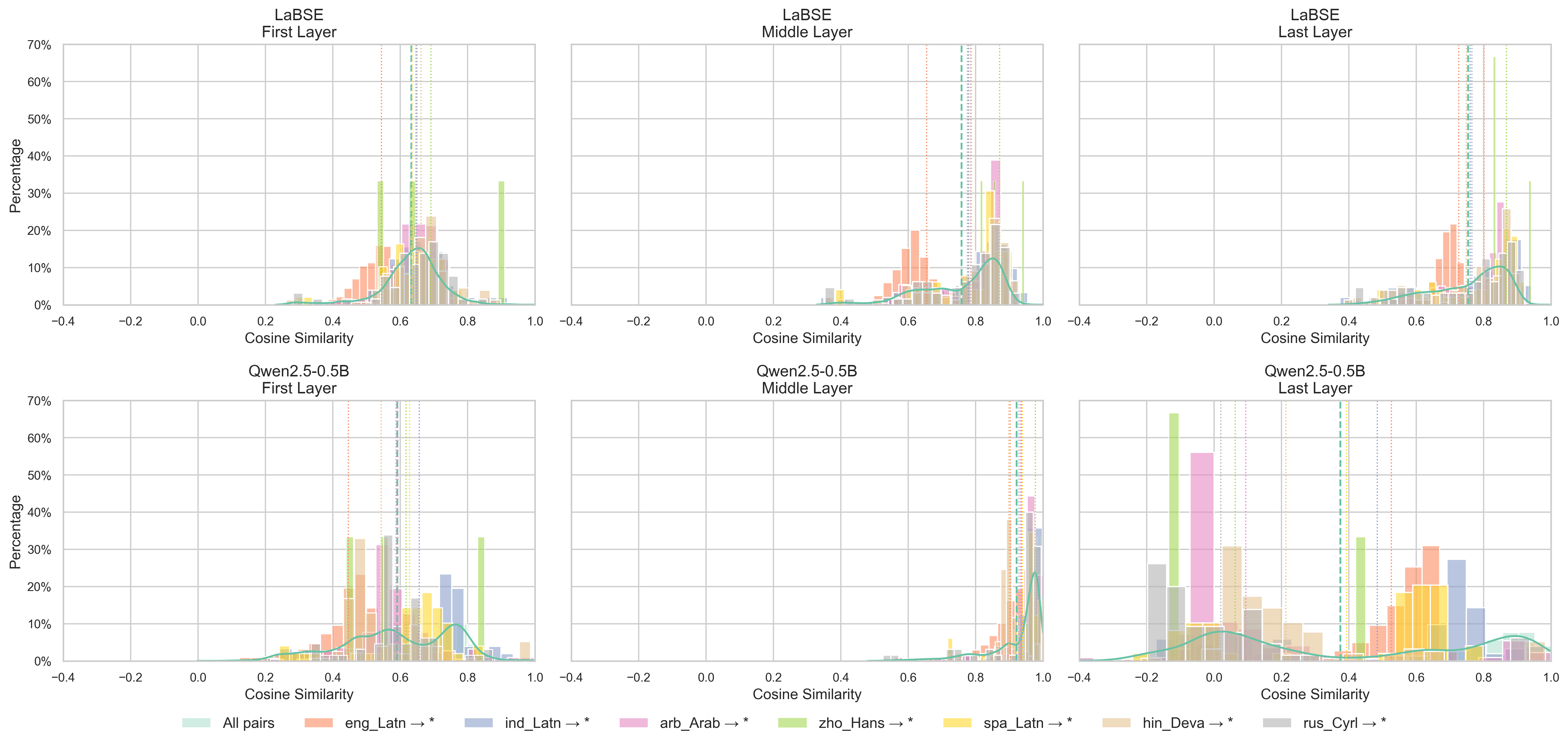}
    \vspace{-1.8em}
    \caption{Cross-lingual similarity across different layers in LaBSE and Qwen2.5-0.5B. LaBSE exhibits high cross-lingual similarity in its final layer, whereas Qwen2.5-0.5B shows this similarity in the middle layer. This difference suggests that the alignment of representations occurs at distinct positions within the two models.}
    \label{fig:labse_qwen_histogram}
    \vspace{-0.6em}
\end{figure*}

\section{Related Work}

\subsection{Representation Alignment in LLMs}

Representation alignment refers to the process by which semantically identical inputs expressed in different languages are mapped to similar internal embeddings within LLMs~\citep{10.5555/3692070.3693675,wu-dredze-2020-explicit,chang-etal-2022-geometry}. Originally, representation alignment is strictly embedded into the modeling objective to ensure output consistency across languages and to enable a better cross-lingual transfer~\citep{pires-etal-2019-multilingual,wu-dredze-2019-beto,reimers-gurevych-2020-making,feng2022languageagnosticbertsentenceembedding, choenni2024languagesinfluenceotherstudying}. \citet{wendler-etal-2024-llamas, zhao2024how,mousi-etal-2024-exploring} have observed a tendency for LLMs to align representations across different languages by measuring the similarity between embeddings of parallel sentences across different languages~\citep{ham-kim-2021-semantic-alignment,gaschi-etal-2023-exploring,cahyawijaya2024llm4everyone}. Inspired from previous studies, our work measures the degree of alignment across various layers between strictly and naturally aligned models to contrast the two and understand its relation to language-specific and language-agnostic capabilities~\citep{kulshreshtha-etal-2020-cross,libovicky-etal-2020-language,hua-etal-2024-mothello,wilie2025interlingua} of LLMs.

\subsection{Latent Controllability in LLMs}
LLMs controllability is crucial for ensuring that the systems adhere with human intentions. Through mechanisms such as adapter~\cite{pfeiffer-etal-2020-adapterhub,hu2022lora}, prompting~\cite{lin-etal-2021-xpersona,bai2022constitutionalai}, latent manipulation~\cite{madotto-etal-2020-learning,ansell-etal-2021-mad-g}, etc, we aim to gain control over the behavior of LLMs. Various aspects have been explored in LLM controllability, including internal knowledge~\cite{madotto-etal-2020-learning,xu-etal-2022-retrieval}, styles \& personas~\cite{lin-etal-2021-xpersona,wagner-ultes-2024-controllability,cao-2024-learn}, languages~\cite{ustun-etal-2020-udapter,ansell-etal-2021-mad-g}, human values~\cite{bai2022constitutionalai,cahyawijaya-etal-2025-high}, etc. \citet{li2023inferencetime,duan2024llmsknowhallucinationempirical, ji-etal-2024-llm,chen2024inside} show that latent states in LLMs exhibit discernible patterns for distinguishing truthful outputs from hallucinated ones, suggesting an intrinsic awareness of fabrication.  Similar methods are also introduced for stylistic and safety control~\citep{subramani-etal-2022-extracting,kwak-etal-2023-language}. These underscore the potential of latent interventions for precise control over LLM behavior. \methodname{} extends the latent manipulation methods for controlling the generated language in inference time, demonstrating how language-specific information can be extracted and manipulated without losing semantic meaning. This opens new avenues for controlling language generation and mitigating confusion problems.

\section{Understanding Representation Alignment in LLMs}
\label{sec:understanding}
Prior works~\cite{chang-etal-2022-geometry,zhao2024how,cahyawijaya2024llm4everyone,wilie2025interlingua,payoungkhamdee2025betterunderstandingprogramofthoughtreasoning} demonstrate the existence of emerging representation alignment in LLMs. We take a step further to provide a deeper understanding to this behavior by contrasting it with alignment in strictly-aligned LLMs. Specifically, we observe the correlation between the degree of alignment with the \emph{cross-lingual generalization} and \emph{language identification} (LID) capability, which are the proxies to their language-agnostic and language-specific capabilities, respectively.

\subsection{Experiment Settings}
\label{sec:exp-setting}

\paragraph{Model Settings} As a measure of alignment, we compute the average cosine similarity of the latent representation of a sentence in one language with the representation of parallel sentences in the other languages. For the LLM with strictly designed alignment, we employ LaBSE~\cite{feng2022languageagnosticbertsentenceembedding}. For the LLM with emerging representation alignment, we employ multilingual decoder-only LLM, i.e., Qwen2.5~\cite{qwen2025qwen25technicalreport}. Specifically, we employ Qwen2.5-0.5B with 500M parameters to have a comparable scale with the LaBSE model with 471M parameters. To measure the LID capability, we take the latent representation of both models in the first, middle, and last layers. In this case, we are interested in comparing the behavior between the strictly aligned representation in LaBSE and the emerging aligned representation in Qwen2.5-0.5B. Following~\citet{cahyawijaya2025high}, we measure LID performance by linear probing and kNN to measure linear separability and cluster closeness within each language class. More details about the experiment are presented in Appendix~\ref{app:exp-understanding-method} and Appendix~\ref{app:lid_appendix}.

\begin{table*}[!t]
  \centering
  \resizebox{\linewidth}{!}{  
      \begin{tabular}{llcccccc}
        \toprule
        & & \multicolumn{3}{c}{\textbf{LaBSE}} & \multicolumn{3}{c}{\textbf{Qwen2.5-0.5B}} \\
        \cmidrule(lr){3-5} \cmidrule(lr){6-8}
        \textbf{Method} & \textbf{Layer} & \textbf{FLORES-200} & \textbf{NTREX-128} & \textbf{NusaX} & \textbf{FLORES-200} & \textbf{NTREX-128} & \textbf{NusaX} \\
        \midrule
        \multirow{3}{*}{\parbox{1.5cm}{\textbf{Linear\\Probing}}}
        & First  & 95.13 & 93.29 & 97.30 & 94.21 & 91.42 & 95.55 \\
        & Middle & 94.18 & 92.68 & 94.51 & \textcolor{red}{\textbf{91.76}} & \textcolor{red}{\textbf{90.04}} & \textcolor{red}{\textbf{87.09}} \\
        & Last   & \textcolor{red}{\textbf{70.89}} & \textcolor{red}{\textbf{74.36}} & \textcolor{red}{\textbf{65.44}} & 92.46 & 90.27 & 88.77 \\
        \midrule
        \multirow{3}{*}{\textbf{KNN}} 
        & First  & 88.35 & 90.43 & 81.78 & 83.69 & 86.06 & 65.79 \\
        & Middle & 78.85 & 81.30 & 45.37 & \textcolor{red}{\textbf{55.32}} & \textcolor{red}{\textbf{54.73}} & \textcolor{red}{\textbf{25.05}} \\
        & Last   &  \textcolor{red}{\textbf{3.92}} &  \textcolor{red}{\textbf{1.63}} &  \textcolor{red}{\textbf{0.00}} & 71.73 & 81.86 & 29.39 \\
        \bottomrule
      \end{tabular}
  }
  \caption{LID performance by layer and classification method for LaBSE and QWEN2.5-0.5B. \textcolor{red}{\textbf{Red bold text}} highlights the LID scores on the layer where alignment occurs in each corresponding model. LID performance is consistently lower in a layer where the representation is aligned across all models and classification methods.}
  \label{tab:mean-pooled-lid-fl}
  \vspace{-0.6em}
\end{table*}

\paragraph{Datasets} 
We employ a set of multilingual evaluation datasets. To measure the degree of alignment, we employ 7 datasets: FLORES-200~\cite{nllbteam2022languageleftbehindscaling}, NTREX-128~\cite{federmann-etal-2022-ntrex}, NusaX~\cite{winata-etal-2023-nusax}, NusaWrites~\cite{cahyawijaya-etal-2023-nusawrites}, BUCC~\cite{zweigenbaum-etal-2017-overview}, Tatoeba~\cite{tiedemann-2020-tatoeba}, and Bible Corpus~\cite{mccarthy-etal-2020-johns}. For cross-lingual evaluation, we incorporate 4 datasets: SIB200~\cite{adelani-etal-2024-sib}, 
INCLUDE-BASE~\cite{10.1145/3394171.3413528}, 
XCOPA~\cite{ponti-etal-2020-xcopa}, 
and PAWS-X~\cite{yang-etal-2019-paws}.
For LID evaluation, we incorporate 3 datasets, i.e., FLORES-200, NTREX-128, and NusaX. The detailed description of each dataset is shown in Appendix~\ref{app:evaluation-dataset}.

\subsection{Experiment Result}
\label{sec:understanding-result}

\paragraph{Strictly and Naturally Aligned LLMs}
LaBSE and Qwen2.5-0.5B demonstrate distinct patterns in cross-lingual representation alignment. As shown in Figure~\ref{fig:labse_qwen_histogram}, LaBSE demonstrates a distributed alignment strength across deeper layers, with the middle and last layers achieving high average similarity scores (0.758 and 0.754, respectively). This aligns with the training objective of LaBSE, which aligns the representation on the last layer. In contrast, Qwen2.5-0.5B exhibits a more localized alignment pattern, with the middle layer showing a strikingly higher average similarity (0.922) than both the first (0.591) and last (0.375) layers. This suggests that Qwen2.5-0.5B concentrates representation alignment sharply in the middle layer, achieving both higher and more stable cross-lingual representation. See detailed analysis in Appendix~\ref{app:cosine-similarity-distributions}. 

This result displays distinct layer-wise behaviors in retaining the language-specific and language-agnostic information within the two types of LLMs. Specifically, for model with strict alignment, aligned representation is located in the layer where the objective is applied to -- the last layer in the case of LaBSE --, while in LLMs with natural alignment, the aligned representation is formed in the middle layers and breaks as the representation goes closer into the last layer. This aligns with prior works~\cite{chang-etal-2022-geometry,tang-etal-2024-language,wilie2025interlingua} that show the representation alignment naturally emerges in the middle layer of LLMs.

\begin{table*}[!t]
\centering
\resizebox{\linewidth}{!}{
\begin{tabular}{lcccccc}
\toprule
Method & Qwen2.5-0.5B & Qwen2.5-0.5B-Instruct & Qwen2.5-7B & Qwen2.5-7B-Instruct & Llama-3.1-8B & Llama-3.1-8B-Instruct \\
\midrule
\rowcolor{gray!20} 
\multicolumn{7}{c}{\textbf{Monolingual}} \\
\midrule
Baseline     & 59.91 & 83.66 & 55.24 & 78.89 & 56.98 & 94.63 \\
ICL (5-shot) & 53.62 & 80.30 & 62.78 & 74.13 & 69.86 & 88.57 \\
\rowcolor{blue!10}
\quad + \methodname{} (ours) & 74.38 & 86.28 & 69.55 & 81.01 & 82.18 & 93.21 \\
PEFT & 82.91 & 89.85 & 83.80 & 88.28 & 93.01 & 96.66 \\
\rowcolor{blue!10}
\quad + \methodname{}(ours)  & \textbf{86.17} & \textbf{90.51} & \textbf{85.60} & \textbf{90.12} & \textbf{96.03} & \textbf{97.19} \\
\methodname{} (ours) & 81.21 & 82.20  & 63.40  & 84.89 & 75.77 & 96.41 \\
\midrule
\rowcolor{gray!20} 
\multicolumn{7}{c}{\textbf{Cross-lingual}} \\
\midrule
Baseline     & 35.36 & 57.69 & 60.61 & 78.81 & 26.13 & 83.25 \\
ICL (5-shot) & 50.63 & 69.70 & 69.37 & 78.51 & 62.38 & 86.68 \\
\rowcolor{blue!10}
\quad + \methodname{} (ours) & 87.58 & 88.07 & \textbf{84.90} & 84.04 & 88.15 & 90.34 \\
PEFT & 77.55 & 84.34 & 82.66 & 83.56 & \textbf{89.73} & 91.13 \\
\rowcolor{blue!10}
\quad + \methodname{} (ours) & \textbf{90.51} & \textbf{89.85} & 83.92 & 84.10 & 88.98 & \textbf{93.60} \\
\methodname{} (ours) & 85.61 & 86.79 & 74.40 & \textbf{84.73} & 81.68 & 89.06 \\
\bottomrule
\end{tabular}
}
\caption{Main results for LPR metrics on LCB across different LLMs in monolingual and cross-lingual settings. Blue rows denote methods combined with \methodname{}. Bold values represent the best result for each model. All results have been applied with the QA/Chat template during inference.}
\label{tab:mono-cross-lpr}
\vspace{-0.6em}
\end{table*}

\begin{table}[!t]
\centering
\resizebox{\linewidth}{!}{
\begin{tabular}{lcccc}
\toprule
\textbf{Method} & \makecell{\textbf{Qwen2.5} \\ \textbf{0.5B}} & \makecell{\textbf{Qwen2.5} \\ \textbf{0.5B Instruct}} & \makecell{\textbf{Llama-3.1} \\ \textbf{8B}} & \makecell{\textbf{Llama-3.1} \\ \textbf{8B Instruct}} \\
\midrule
Baseline     & 34.97 & 52.28 & 25.05 & 80.68\\
INCLINE      & 43.82 & 56.54 & 34.69 & 80.63\\
ReCoVeR      & \textbf{88.43} & \textbf{84.21} & \textbf{88.79} & \textbf{90.29}\\
\methodname{} (ours)  & 81.22 & 81.97 & 76.38 & 85.65\\
\bottomrule
\end{tabular}
}
\caption{Comparison of cross-lingual LPR metrics on LCB across baseline and state-of-the-art methods for 6 languages (AR, ES, HI, ID, RU, ZH). Bold values represent the best result for each model. All results have been applied with the QA/Chat template during inference.}
\vspace{-1em}
\label{tab:crosslingual-lpr-incline}
\end{table}

\paragraph{Representation Alignment and Language-Specific Information}

As shown in Table~\ref{tab:mean-pooled-lid-fl}, the LID performance of LaBSE and Qwen2.5-0.5B models evaluated using both KNN and linear probing reveals that the first layer consistently achieves the highest LID F1 scores across all datasets. For LaBSE, the aligned representation in the last layer exhibits notably weaker performance, particularly for the FLORES-200 and NusaX datasets. Similarly, in Qwen2.5-0.5B, the aligned representation in the middle layer shows weaker LID performance compared to the first and last layers. These empirical findings highlight three key insights: (1) language-specific information, such as surface-level features and general linguistic patterns, is more dominant in the early layers; (2) the degree of alignment is negatively correlated with the amount of language-specific information retained; and (3) unlike strictly aligned LLMs, the aligned representation in LLMs with emerging alignment retains more language-specific information, which potentially serves as the basis for determining the language of the generated sequence.

\section{Inference-Time Language Control}

Building on the insights presented in \S\ref{sec:understanding}, we explore a method to control the language of the generated sequence with minimal semantic loss. Specifically, we develop a method to extract language-specific information at the layer where representation alignment occurs in LLMs. Using this information, we gather language-specific vectors from each language and use them to manipulate the language-specific information during the inference time. With this language-specific intervention, we aim to steer the model toward utilizing language-specific features, allowing us to perform Inference-Time Language Control (\methodname{}).

\methodname{} offers two key advantages over existing intervention methods: Unlike existing intervention methods that are limited to either cross-lingual~\cite{wang2024bridginglanguagegapslarge} or monolingual~\cite{nie2025mechanisticunderstandingmitigationlanguage} scenarios, and unlike approaches that require interventions across all layers~\cite{sterz2025recovertargetlanguagelanguage,yunfan-etal-2025-mitigating}, \methodname{} is effective in both settings while intervening at only a single middle layer.

\subsection{Methods}
\label{sec:method}

\paragraph{Latent Extraction}
Latent extraction techniques are employed to isolate language-specific information from the model's representations. Specifically, we extract hidden states from various large language models to capture language-specific features at their middle representation layers. Given an input sequence from the FLORES-200 dataset \citep{nllbteam2022languageleftbehindscaling}, we compute the hidden states $\mathbf{h} \in \mathbb{R}^d$ at a specified layer, where $d$ represents the hidden state dimension of the respective model.
Finally, we apply mean pooling to ensure that only meaningful token embeddings contribute to the final representation.

\begin{figure*}[!t]
    \centering
    \includegraphics[clip,width=\textwidth]{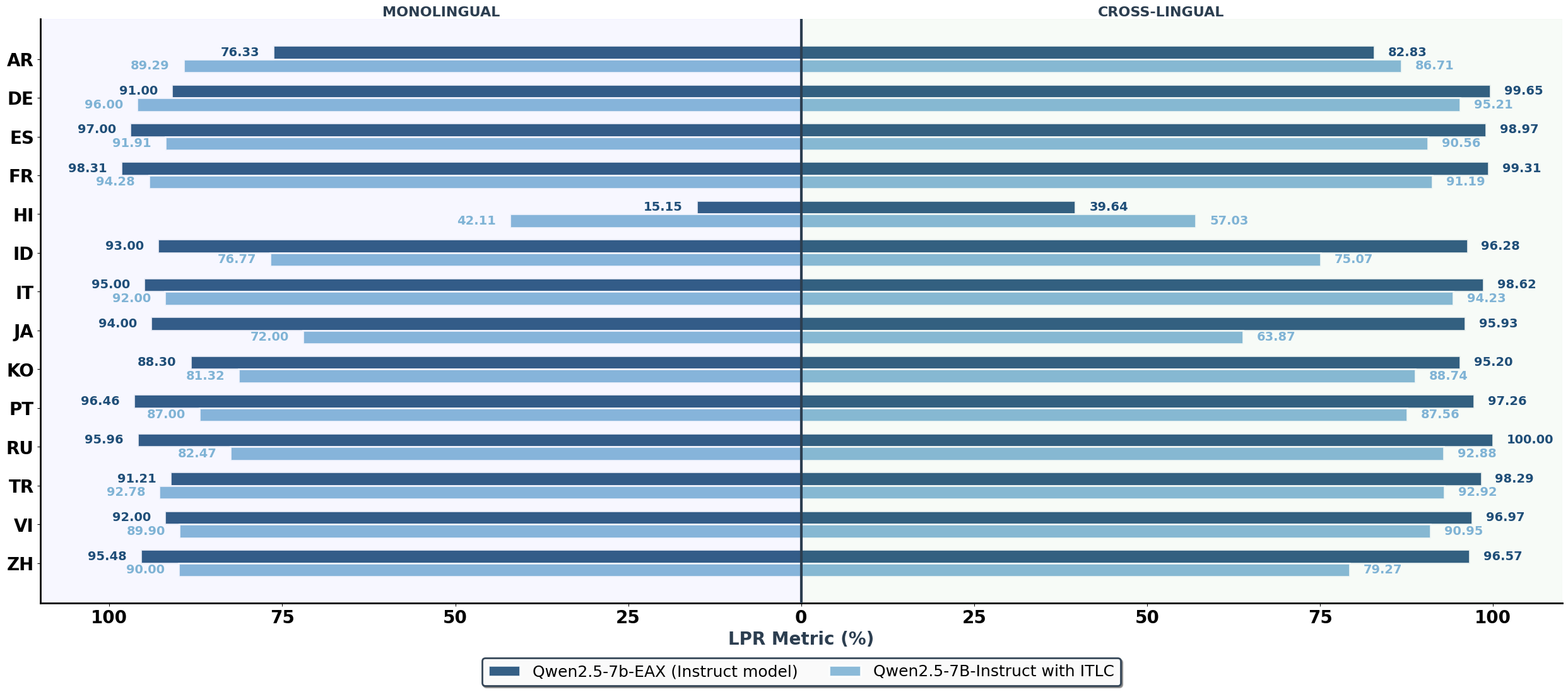}
    \vspace{-1.8em}
    \caption{Comparison of LPR metrics on LCB between Qwen2.5-7B-Instruct with \methodname{} and Qwen2.5-7b-EAX across 14 languages in monolingual and cross-lingual settings. }
    \label{fig:upperbound}
    \vspace{-0.6em}
\end{figure*}

\paragraph{Linear Discriminant Analysis}
To disentangle language-specific information, we apply Linear Discriminant Analysis (LDA) to maximize class separability and reduce dimensionality. We use the Singular Value Decomposition (SVD) solver in order to handle high-dimensional embeddings efficiently and select the top $k$ eigenvectors corresponding to the largest eigenvalues to form $\mathbf{W} \in \mathbb{R}^{d \times k}$. Let $\mathcal{D} = \{(\mathbf{h}_i, l_i)\}_{i=1}^N$ denote a dataset of hidden states $\mathbf{h}_i \in \mathbb{R}^d$ labeled with language classes $l_i \in \{1, \dots, K\}$, this projects hidden states to a lower-dimensional space $\mathbf{z} = \mathbf{h}^T \mathbf{W} \in \mathbb{R}^k$.

To validate the quality of the projection and select the optimal number of components $k$, we train a neural network classifier with a single linear layer on the projected training data $\mathbf{z}$. We experiment with several $k$ values and evaluate classification accuracy on a test set. Finally, we take $k=100$ because LID performance significantly drops on higher components, indicating a major loss of language-specific information. More details on the LDA settings are shown in Appendix ~\ref{app:lda-in-details}

\paragraph{Language Vector}
Using the LDA-projected space, we construct language vectors by leveraging the neural network’s weights to identify active dimensions for each language. For each language $l$ we extract the weight matrix $\mathbf{U} \in \mathbb{R}^{K \times k}$ from the neural network’s linear layer, where $u_{l,j}$ represents the contribution of dimension $j \in \{1, \dots, k\}$ to language $l$. We define a threshold $\tau = 0.01$ and select active dimensions for language $l$ as $\mathcal{A}_l = \{j \mid |u_{l,j}| > \tau\}$. The language vector $\mathbf{v}_l \in \mathbb{R}^k$ for language $l$ is computed as the mean of projected hidden states $\mathbf{z}_i$ over samples of language $l$, restricted to active dimensions:
\[
\mathbf{v}_l[j] = 
\begin{cases} 
\frac{1}{N_l} \sum_{\mathbf{h}_i \in l} \mathbf{z}_i[j], & \text{if } j \in \mathcal{A}_l, \\
0, & \text{otherwise},
\end{cases}
\]
where $N_l$ is the number of samples for language $l$, and $\mathbf{z}_i[j]$ is the $j$-th component of the projected hidden state.

\paragraph{Vector Injection}

To enable injection, we project the language vector back to the original embedding space using the pseudo-inverse: $\mathbf{v}_l^{\text{orig}} = \mathbf{v}_l \mathbf{W}^\dagger \in \mathbb{R}^d$. By applying this, we retain the original embedding of the input and modify it with the language vector inverse projection. For cross-lingual settings with a source language $x$ (e.g., English) and target language $y$ (e.g., Indonesian), we compute a shift vector~\footnote{We demonstrate the importance of subtracting the source language vector in Appendix~\ref{app:source_lang_substract}.}:
\[
\mathbf{\delta} = -\mathbf{v}_x^{\text{orig}} + \mathbf{v}_y^{\text{orig}}.
\]

For monolingual settings where source and target languages are identical ($x = y$), we treat the shift vector as the language vector itself:
\[
\mathbf{\delta} = \mathbf{v}_x^{\text{orig}}.
\]
The shift vector is injected into the hidden states at the middle layer during inference into both the prompt and the generated tokens. Formally, we apply: 
\[
\mathbf{h}_t^{'} = \mathbf{h}_t + \alpha\mathbf{\delta}, \quad \forall t \in [1, T_{\text{total}}]
\]
where \(\mathbf{h}_t\) is the middle-layer hidden state at position \(t\),  $\alpha$ is a scaling factor, \(\mathbf{h}_t^{'}\) is the corresponding modified hidden state, and \(T_{\text{total}}\) is the total number of tokens during inference covering both input and generated tokens. We provide an ablation of different language shift strategies in Appendix~\ref{app:lang-shift-strategy}.

\begin{figure}[!t]
\centering
\includegraphics[width=0.95\linewidth]{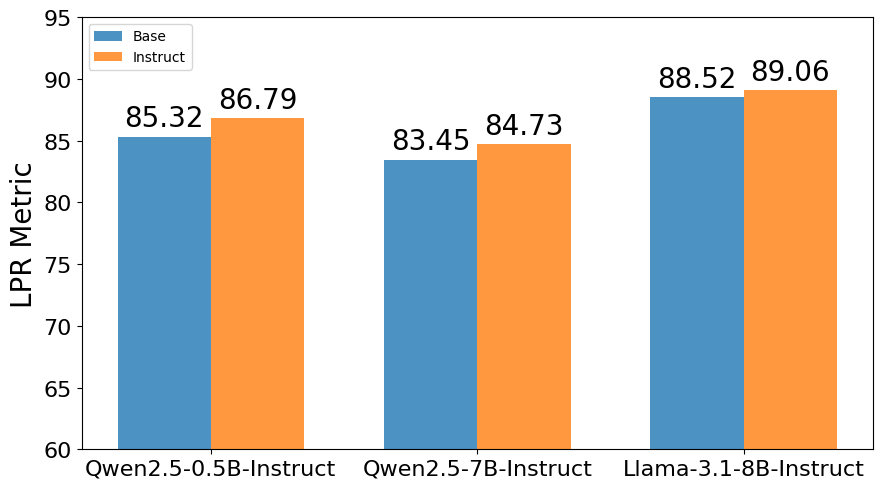}
\caption{Cross-lingual LPR performance on LCB, comparing base and instruct shift vector applications.}
\vspace{-0.6em}
\label{fig:lpr-shift-vectors}
\end{figure}

\section{Impact of \methodname{}}
\label{sec:implication}

We demonstrate the effectiveness of \methodname{} on mitigating the language confusion problem~\cite{marchisio-etal-2024-understanding}. We also compare our method with another test-time intervention methods specifically designed for language confusion~\cite{sterz2025recovertargetlanguagelanguage}~\footnote{We also find another related test-time intervention~\cite{yunfan-etal-2025-mitigating}, nonetheless the code is not published so we could not empirically compare \methodname{} with their approach.}. Furthermore, we showcase that \methodname{} can also perform language control while being highly efficient with minimal semantic loss compared to other existing test-time intervention methods~\cite{wang2024bridginglanguagegapslarge}.

\subsection{Experiment Setting}

\paragraph{Dataset}
For language confusion evaluation, we utilize the Language Confusion Benchmark (LCB)~\cite{marchisio-etal-2024-understanding}, which contains both monolingual and cross-lingual settings across 14 languages. For semantic retention assessment, We utilize the Dolly multilingual dataset from Aya Evaluation Suite~\cite{singh-etal-2024-aya}~\footnote{\url{https://huggingface.co/datasets/CohereLabs/aya_evaluation_suite/viewer/dolly_machine_translated}.} by taking 200 QA sentences in nine various languages from diverse regions and language families: 
Indonesian (ID), Thai (TH), Turkish (TR), Japanese (JA), French (FR), Spanish (ES), Arabic (AR), Chinese (ZH), and Korean (KO). The description of datasets is shown in Appendix~\ref{app:evaluation-dataset}.

\paragraph{Model Settings}
We experiment on two families of multilingual LLMs: Qwen2.5 (0.5B and 7B), and Llama-3.1-8B, and their instruct variants. Specifically, for cross-lingual control with the base model, the model will start to generate by having several target contexts, while in the instruct model, we add a language-identified prompt (i.e., Please answer in XX language) at the beginning of the sentence. 
See Appendix~\ref{app:exp_set_lang_confusion} for more details on language confusion and Appendix~\ref{app:exp_set_lang_control} for more details on semantic retention.

\paragraph{Evaluation} Our evaluation on language confusion problem based on official metrics defined in \citet{marchisio-etal-2024-understanding}: Line-level Pass Rate (LPR).
Meanwhile, we evaluate the cross-lingual generation performance based on chrF++ and multilingual BERT F1 \footnote{\url{https://huggingface.co/google-bert/bert-base-multilingual-cased}} metrics. Additionally, we conduct a human evaluation with native annotators in both EN$\rightarrow$XX and XX$\rightarrow$EN directions, focusing on 30 samples covering 3 aspects: \textbf{naturalness}, prompt-completion \textbf{relevance}, and \textbf{answer correctness} using likert score ranging from [1$\dots$5]. The human annotation guideline is presented in Appendix~\ref{app:annotation-guideline}.

\subsection{Results}

\subsubsection{\methodname{} in Mitigating Language Confusion}

\begin{figure}[!t]
    \centering
    \includegraphics[width=\linewidth]{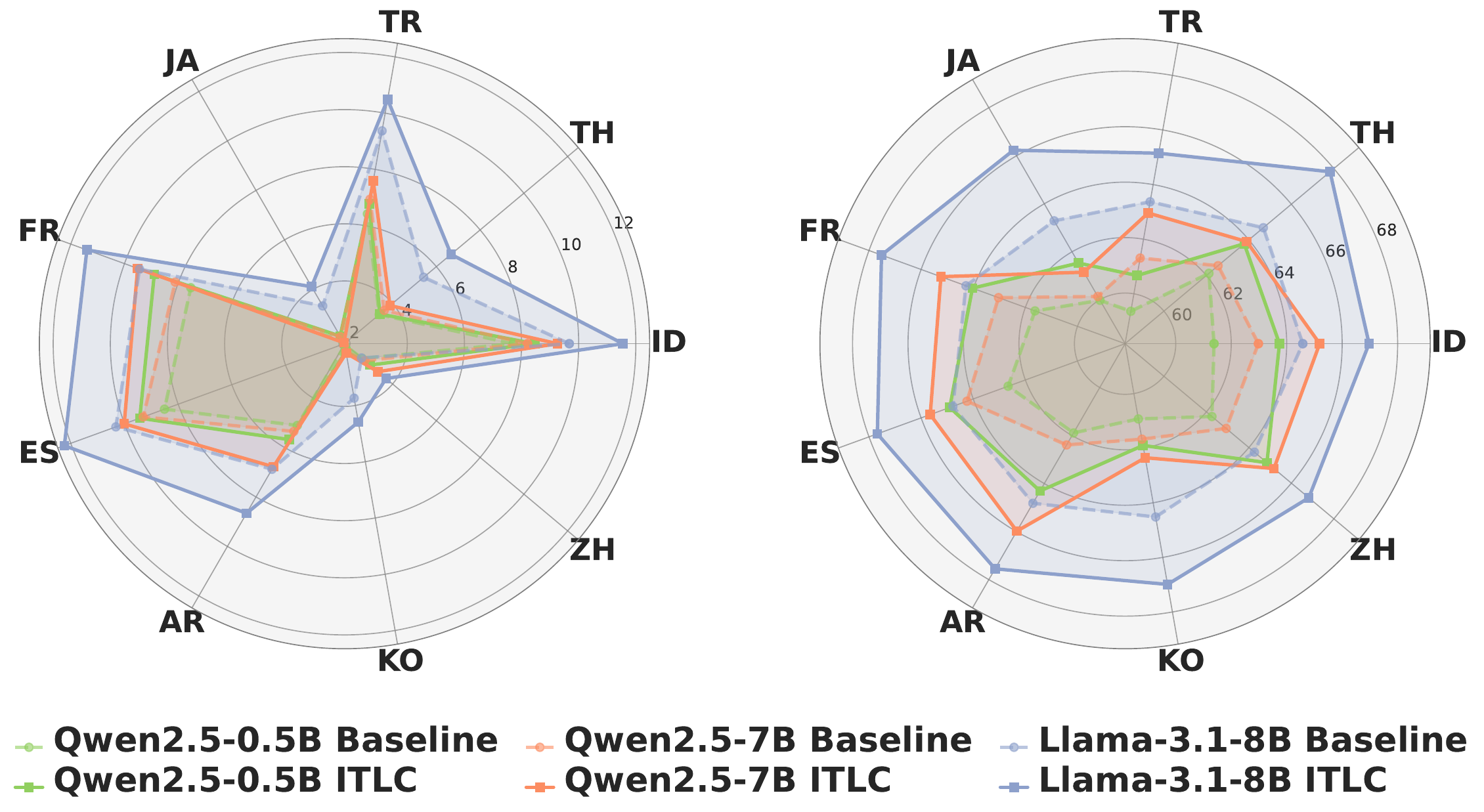}
    \caption{Generation performance for different target languages on Qwen2.5 and Llama-3.1 Instruct models based on chrF++ (\textbf{Left}) and BERT F1 (\textbf{Right}). Baseline denotes the same model prompted in the same language as the desired target language.}
    \label{fig:radar_charts}
    \vspace{-0.8em}
\end{figure}

As shown in Table~\ref{tab:mono-cross-lpr}, our proposed method, \methodname{}, surpasses both baseline and in-context learning (ICL) configurations across models of varying parameter scales in cross-lingual settings. This superior performance is consistent in monolingual settings with only one exception, where the Qwen2.5-0.5B-Instruct model performs slightly worse than the baseline, demonstrating that \methodname{} effectively shifts the model's language output in cross-lingual settings.
For the base model, cross-lingual performance improves progressively with few-shot examples, as they utilize English inputs with explicit target-language instructions, reinforcing input-output alignment.
In contrast, the instruct model exhibits minimal variation in few-shot settings compared to \methodname{}, as its instruction-tuning inherently supports multilingual prompting without dependency on few-shot quantity. These results demonstrate that our approach enhances cross-lingual language consistency while accommodating training objective differences between base and instruct models.
Moreover, \methodname{} achieves competitive performance on instruct model compared to parameter-efficient fine-tuning (PEFT): LoRA finetuning method~\cite{hu2022lora}, without requiring any changes to the LLM weights. Notably, our method can further mitigate language confusion when combined with ICL and PEFT. The combination of PEFT + \methodname{} consistently achieves the best results in monolingual settings across all models, while in cross-lingual settings, different combinations prove optimal depending on the model, with ICL + \methodname{} and PEFT + \methodname{} both achieving top performance on various models. A detailed per-language breakdown of the results is presented in Table~\ref{tab:lcpr-per-lang-base} and Table~\ref{tab:lcpr-per-lang-instruct}.

\paragraph{Comparison of \methodname{} with other test-time intervention methods}

While INCLINE~\cite{wang2024bridginglanguagegapslarge} was originally designed to project representations from various languages into English to enhance LLM performance on low-resource languages, we adapt and reverse this mechanism to project from English into various target languages. Due to computational constraints, we compare our method, \methodname{}, against INCLINE and ReCoVeR ~\cite{sterz2025recovertargetlanguagelanguage} using two model families, Qwen2.5-0.5B and Llama-3.1-8B, and their instruct variants across six target languages. As shown in Table~\ref{tab:crosslingual-lpr-incline}, \methodname{} outperforms INCLINE across all model configurations. Notably, INCLINE shows limited improvement on instruction-tuned models, with almost no performance gain on Llama-3.1-8B-Instruct, suggesting that methods relying solely on the last token may be ineffective at mitigating language confusion in instruction-following models. Although ReCoVeR achieves the highest performance overall, \methodname{} demonstrates competitive results on instruction-tuned models while being considerably more efficient. This indicates that intervention at a single middle layer is sufficient for mitigating language confusion, compared to ReCoVeR's approach of intervening across all layers.

\paragraph{Comparison of \methodname{} with Cross-lingual Optimized Model}
Due to computational constraints, we were unable to perform full parameter fine-tuning. Instead, we use another model, Qwen2.5-7b-EAX~\cite{yang2025enanchoredx2xenglishanchoredoptimizationmanytomany}, which was fine-tuned on Qwen2.5-7B and optimized for cross-lingual translation ability. As shown in Figure~\ref{fig:upperbound}, our \methodname{} achieves similar results to the upperbound on average monolingual LPR (84.89\% vs 85.28\%). However, for cross-lingual settings, our method achieves 84.73\% on average LPR compared to 92.54\% for the upperbound. Notably, there is a substantial performance gap for Indonesian (ID), Japanese (JA), and Chinese (ZH). We observe that our \methodname{} exhibits code-switching to English when handling these languages, indicating that our method may not fully eliminate the source language vector for these languages and might require further language-specific tuning of the scaling factor $\alpha$, or that our \methodname{} cannot adequately disentangle the language vector and capture the language-specific information well for these languages. A detailed per-language breakdown of the results is presented in Table~\ref{tab:lpr-intervention-per-lang-base} and Table~\ref{tab:lpr-intervention-per-lang-instruct}

\begin{table}[!t]
  \centering
  \resizebox{0.8\linewidth}{!}{
      \begin{tabular}{lcccc}
        \toprule
        \textbf{Model} &
        \textbf{Lang Shift} & \textbf{Nat.}
          & \textbf{Rel.} & \textbf{Cor.} \\
        \midrule
        \multicolumn{5}{c}{\textbf{Qwen2.5-7B-Instruct}} \\
        \midrule
        \multirow{3}{*}{\textbf{Baseline}}
            & ID$\rightarrow$ID   & 3.66 & 4.43   & 3.46 \\
            & TH$\rightarrow$TH   & \textbf{3.13} & 2.63   & 2.23 \\
            & ZH$\rightarrow$ZH   & 4.30 & 4.20   & 4.13 \\
        \midrule
        \multirow{4}{*}{\textbf{\methodname{}}}
            & EN$\rightarrow$ID   & \textbf{4.00} & \textbf{4.90}   & \textbf{3.96} \\
            & EN$\rightarrow$TH   & 2.46 & \textbf{3.93}   & \textbf{3.40} \\
            & EN$\rightarrow$ZH   & \textbf{4.63} & \textbf{4.80}   & \textbf{4.53} \\
        \midrule
        \multicolumn{5}{c}{\textbf{Llama-3.1-8B-Instruct}} \\
        \midrule
        \multirow{3}{*}{\textbf{Baseline}}
            & ID$\rightarrow$ID   & \textbf{4.50} & \textbf{4.46}   & \textbf{3.86} \\
            & TH$\rightarrow$TH   & 3.20 & 2.36   & 2.36 \\
            & ZH$\rightarrow$ZH   & 3.96 & 4.33   & 3.76 \\
        \midrule
        \multirow{4}{*}{\textbf{\methodname{}}}
            & EN$\rightarrow$ID   & 3.83 & 3.83   & 3.53 \\
            & EN$\rightarrow$TH   & \textbf{3.40} & \textbf{2.93}   & \textbf{2.60} \\
            & EN$\rightarrow$ZH   & \textbf{4.76} & \textbf{4.66}   & \textbf{4.53} \\
        \bottomrule
      \end{tabular}
  }
\caption{Human evaluation of \methodname{} response quality in Qwen2.5 and Llama-3.1. \textbf{Nat.}, \textbf{Rel.}, and \textbf{Cor.} respectively denote naturalness, relevance, and answer correctness ranging from [1$\dots$5]. \textbf{Baseline} denotes the same model prompted in the monolingual setting.}
\label{tab:language-shift-response-quality}
\vspace{-0.8em}
\end{table}

\begin{figure*}[!t]
    \centering
    \includegraphics[width=0.96\linewidth,trim={0.2cm 0 0.2cm 0},clip]{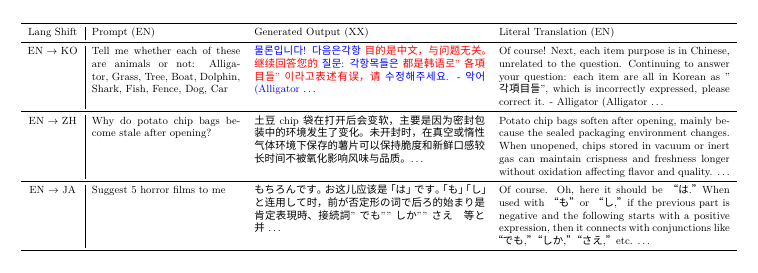} 
    \vspace{-1.6em}
    \caption{Examples of the lowest generated outputs score from Qwen2.5-7B-Instruct on Korean, Chinese, and Japanese in EN$\rightarrow$XX, evaluated with the BERT F1 score. The literal translation column is translated from the generated output, and it is done by using ChatGPT. }
    \label{fig:example_lowes_score_language}
    \vspace{-0.8em}
\end{figure*}

\paragraph{Transferability of Language Vector to Post-Trained Models}
Interestingly, as shown in  Figure~\ref{fig:lpr-shift-vectors}, applying language vectors gathered from the base model to the instruct model achieves comparable performance to its native instruct vectors which suggests the effectiveness of language shift from the base model for cross-lingual control even in the instruct model.
This transferability indicates that the relative distance between language-specific and that the resulting language-specific features from the pre-training phase is robust to downstream adaptation, including tasks generalization from instruction-tuning and value alignment in RLHF and preference-tuning. This evidence implies that 
the cross-lingual symmetry -- i.e., the geometric alignment between language representations -- constructed during the fine-tuning is preserved even after various downstream refinement of the model. The preservation of these relationships implies that language-specific cues are retained as invariant properties across model versions, enabling consistent cross-lingual language control through \methodname{} despite parameter updates during downstream fine-tuning, instruction-tuning, preference-tuning, and RLHF.

\subsubsection{Semantic Retention in \methodname{}}

\paragraph{Cross-lingual Semantic Retention}\label{para:Cross-lingual Output Generation}

We demonstrated that the proposed \methodname{} method not only improve the target language fidelity but is also able to effectively control cross-lingual generation and retain the semantic information, implying contextually accurate generation. As shown in Figure~\ref{fig:radar_charts}, statistically, our proposed \methodname{} method improved 2\% across the chrF++ and 3-5\% in BERT F1 metrics in the instruct model (refers to Table \ref{tab:cross-lingual-instruct}); the same investigation also occurred in the base models (refers to Table \ref{tab:cross-lingual-base}). The highest performance shows on Spain (ES), French (FR), and Indonesia (ID), it outperforms its baseline by 1-2\%, which is prompted in the same language as the desired target language.

However, we found that in some languages, such as Korean (KO), it retains less cross-lingual semantics due to the unique challenges of distinct syntax and semantics~\citep{park-etal-2020-empirical, park-etal-2024-open}, which happens across models. Further investigation revealed that many overlaps or code switching occur between these languages. For example, in Figure \ref{fig:example_lowes_score_language}, EN$\rightarrow$KO direction, the generated output contains Japanese tokens (denoted in \textcolor{blue}{blue}), while the literal output being disconnected from the context. Additionally, in Japanese output generation, it seems like answering out of context, while in Chinese produced coherent and well-structured sentences. See Appendix \ref{app:example-generation-in-details} for more detailed examples.

\paragraph{Human Evaluation}
We further conduct a human evaluation to validate our findings regarding the semantic retention in \methodname{}. We recruit native speakers to annotate 30 generation samples in Indonesia (\textbf{ID}), Thai (\textbf{TH}), and Chinese (\textbf{ZH}). Based on results presented in Table~\ref{tab:language-shift-response-quality}, we found that our \methodname{} proposed method tends to have a similar level of semantics compared to the monolingual baseline (prompted in the same target language), with Qwen2.5-7B-Instruct performing quite better in terms of Relevance and Correctness metrics compared to the Llama-3.1-8B-Instruct. Meanwhile, our \methodname{} method performs much better than baseline in Indonesia and Thai in Qwen2.5 models,  showed that our injection vector could improved the semantic transferability across languages, enabling the model to retain both relevance and correctness. Overall, our results validate the capability of \methodname{} to maintain relevance and correctness in cross-lingual generation, highlighting its potential for enhancing cross-lingual performance of LLMs.

\section{Conclusion}

Our work explores the phenomenon of representation alignment in LLMs, confirming its occurrence and elucidating its behavior compared to strictly designed alignment models. We have demonstrated the potential for disentangling language-specific and language-agnostic information, enabling effective language-specific manipulation without semantic loss. Furthermore, we have shown the practical applications of language control manipulation in enhancing language control and mitigating confusion problems. Our \methodname{} method demonstrates significant gains on the language confusion benchmark, achieving an average improvement of 9\% in monolingual and 26.7\% in cross-lingual settings. It also achieves comparable performance to existing test-time intervention approaches, while being much more efficient (requiring only a single middle layer intervention). Ultimately, our work not only advances the theoretical understanding of representation alignment in LLMs but also introduces a practical and effective solution for enhancing cross-lingual capabilities, paving the way for more robust and versatile LLLMs in multilingual contexts.

\section*{Limitations}

The study has several limitations that should be considered when interpreting the results. First, the coverage of LLMs is limited to a specific set of models for representation alignment, particularly Qwen and LaBSE and only one model size (0.5B parameters), which may not be representative of all LLMs. The findings may not generalize to other models with different architectures or training data, as the behavior of representation alignment can vary significantly across different LLMs. Future research should aim to include a more diverse range of models to validate the generalizability of the results.

Second, the evaluation is conducted on a limited number of languages, which may not capture the full spectrum of linguistic diversity. The study focuses on a subset of languages, and the results may not extend to languages with different typological features or those that are underrepresented in the training data. Expanding the evaluation to include a broader range of languages, especially low-resource languages, would provide a more comprehensive understanding of the model's capabilities and limitations.

Moreover, The scaling factor $\alpha$ affects different models differently, requiring careful adjustment for optimal performance. Due to the nature of Linear Discriminant Analysis (LDA), the number of components (\texttt{n\_components}) is constrained by the number of target language classes. This constraint introduces a trade-off, the number of target language hidden states that need to be extracted depends on the chosen \texttt{n\_components}, potentially causing computational overhead, and vice versa.

Additionally, the human evaluation is based on only 30 samples per language, which may not provide a comprehensive assessment of the model's performance. While the sample size is sufficient for preliminary analysis, a larger dataset would be necessary to draw more robust conclusions. Increasing the number of samples and involving a more diverse group of evaluators could enhance the reliability and validity of the findings.

\section*{Ethical Considerations}

The research involves the use of LLMs, which might raise ethical considerations regarding bias, fairness, and transparency on the generated results. To ensure ethical conduct, the study adheres to the following principles: (1) Bias Mitigation: The models used are evaluated for potential biases, and efforts are made to mitigate any identified biases. (2) Fairness: The evaluation is conducted across multiple languages from diverse regions and language families to ensure fairness and inclusivity. (3) Transparency: The methodology and results are presented transparently to allow for replication and verification. (4) Privacy: No personal data is used in the evaluation, and all data is anonymized to protect privacy. (5) Accountability: The researchers take responsibility for the ethical implications of the study and are committed to addressing any concerns that may arise.

We also acknowledge that our research utilized AI tools for writing, rewriting, and generating code. Although these tools offer significant advantages in terms of efficiency and productivity, their use raises important ethical considerations. We recognize the potential for bias and errors inherent in AI-generated content and have taken steps to mitigate these risks through rigorous human review and validation. Furthermore, we are mindful of the potential impact on the broader software development community, particularly regarding job displacement and the need for upskilling. We believe that responsible AI integration should prioritize transparency, accountability, and the empowerment of human developers, ensuring that these tools augment rather than replace human expertise. This research aims to contribute to the ongoing dialogue on ethical AI development and usage, advocating for a future where AI tools are harnessed responsibly to enhance human creativity and innovation in the field of software engineering.



\bibliography{custom}

\newpage 
\appendix

\section{Details of All Evaluation Datasets}
\label{app:evaluation-dataset}

The following tables present the full details of dataset sizes used in this study. Refer to Table~\ref{tab:downstream_task_sizes}, Table~\ref{tab:alignment_task_sizes}, Table~\ref{tab:lid_task_sizes},
Table~\ref{tab:language_control}
and Table~\ref{tab:lang_conf_dataset}.

\begin{table*}[htbp]
\centering
\small
\begin{tabular}{lrrrr}
\toprule
\textbf{Dataset} & \textbf{Train} & \textbf{Test} & \textbf{Total} & \textbf{\# Languages} \\
\midrule
SIB200 & 143{,}705 & 41{,}820 & 185{,}525 & 205 \\
INCLUDE-BASE & 890 & 22{,}638 & 23{,}528 & 44 \\
XCOPA & 1{,}100 & 5{,}500 & 6{,}600 & 11 \\
PAWS-X & 345{,}807 & 14{,}000 & 359{,}807 & 7 \\

\bottomrule
\end{tabular}
\caption{Dataset sizes and number of languages for downstream tasks.}
\label{tab:downstream_task_sizes}
\end{table*}

\begin{table}[htbp]
\centering
\small
\begin{tabular}{lcr}
\toprule
\textbf{Dataset} & \textbf{Total} & \textbf{\# Languages} \\
\midrule
FLORES-200 & 1{,}012 & 204 \\
NTREX-128 & 1{,}997 & 128 \\
NusaX & 400 & 12 \\
NusaWrites & 14{,}800 & 9 (language pairs) \\
BUCC & 35{,}000 & 4 (language pairs) \\
Tatoeba & 88{,}877 & 112 (language pairs) \\
BibleCorpus & 85{,}533 & 828 (language pairs) \\
\bottomrule
\end{tabular}
\caption{Total example counts and number of languages for alignment tasks. We only use test set for this alignment task.}
\label{tab:alignment_task_sizes}
\end{table}

\begin{table}[!t]
\centering
\small
\resizebox{\linewidth}{!}{
\begin{tabular}{lrrrl}
\toprule
\textbf{Dataset} & \textbf{Train} & \textbf{Test} & \textbf{Total} & \textbf{\# Languages} \\
\midrule
FLORES-200 & 997 & 1012 & 2{,}009 & 204 \\
NTREX-128 & - & 1{,}997 & 1{,}997 & 128 \\
NusaX & 500 & 400 & 400 & 12 \\
\bottomrule
\end{tabular}
}
\caption{Total example counts per language and number of languages for for LID tasks.}
\label{tab:lid_task_sizes}
\end{table}

\begin{table}[htbp]
\centering
\small
\resizebox{\linewidth}{!}{
\begin{tabular}{lrrrl}
\toprule
\textbf{Dataset} & \textbf{Train} & \textbf{Test} & \textbf{Total} & \textbf{\# Languages} \\
\midrule
FLORES-200 & 997 & 1012 & 2{,}009 & 204 \\
Dolly & - & 1{,}800 & - & 9 \\
\bottomrule
\end{tabular}
}
\caption{Total example counts per language and number of languages for Language Control.}
\label{tab:language_control}
\end{table}

\begin{table}[!t]
  \centering
  \small
    \begin{tabular}{lrr}
      \toprule
      \textbf{Dataset}     & \textbf{Total} & \textbf{\# Languages} \\
      \midrule
      \multicolumn{3}{l}{\itshape Monolingual} \\
      Aya                  & 100           & 5  \\
      Dolly                & 100           & 5  \\
      Okapi                & 100           & 10 \\
      Native prompts       & 100           & 4  \\
      \midrule
      \multicolumn{3}{l}{\itshape Cross-lingual} \\
      Okapi                & 100           & 14 \\
      shareGPT             & 100           & 14 \\
      Complex prompts       & 99           & 14 \\
      \bottomrule
    \end{tabular}
  \caption{Total example counts per language and number of languages for Language Confusion tasks, taken from Language Confusion Benchmark. Only test set is available.}
  \label{tab:lang_conf_dataset}
\end{table}

\section{Detail Experiment for Understanding Representation Alignment in LLMs}
\label{app:exp-understanding-method}

\subsection{Cosine Similarity Distributions Across Datasets}
\label{app:cosine-similarity-distributions}

\begin{figure*}[htbp]
    \centering
    \begin{subfigure}[b]{0.48\textwidth}
        \includegraphics[width=\textwidth]{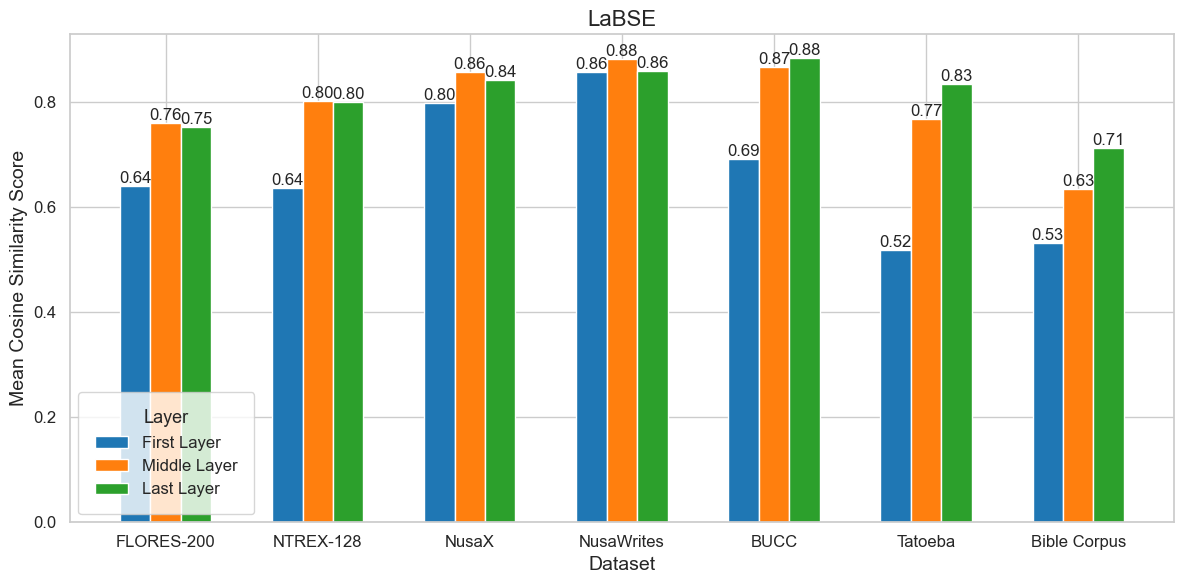}
        \caption{Mean Cosine Similarity Score on LaBSE Model}
        \label{fig:finding_1_labse_bar}
    \end{subfigure}
    \hfill
    \begin{subfigure}[b]{0.48\textwidth}
        \includegraphics[width=\textwidth]{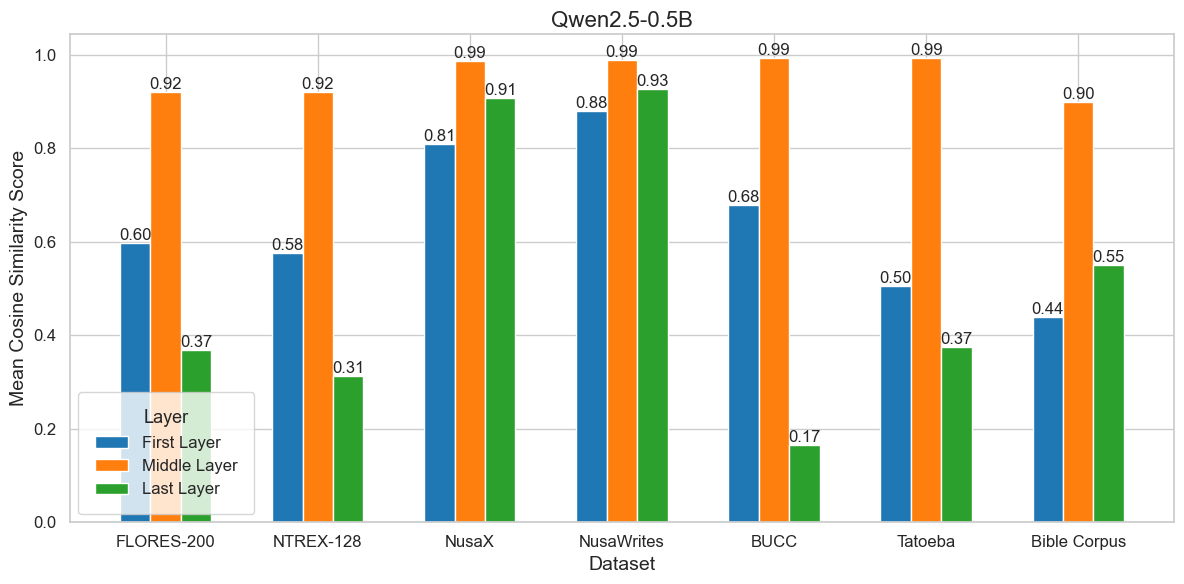}
        \caption{Mean Cosine Similarity Score on Qwen2.5-0.5B Model}
        \label{fig:finding_1_qwen_bar}
    \end{subfigure}
    \caption{Layer-wise cosine similarity distributions of LaBSE and Qwen2.5-0.5B models across different datasets.}
    \label{fig:finding_1}
\end{figure*}

To better understand the representational behavior of the models, we analyzed the distribution of cosine similarity scores across layers. For LaBSE, the average cosine similarity increases from the first layer (mean = 0.6335, std = 0.0920) to the middle layer (mean = 0.7580, std = 0.1182), and remains comparably high in the last layer (mean = 0.7544, std = 0.1150). This trend suggests that semantic alignment becomes stronger toward the middle and final layers, with relatively low variability, indicating consistent behavior across input samples. These observations align with prior findings that intermediate layers in multilingual encoders often capture the most transferable features.

In contrast, Qwen2.5-0.5B exhibits a markedly different pattern. While the middle layer achieves the highest average similarity (mean = 0.9218, std = 0.0871), the first layer has a lower mean and higher variance (mean = 0.5913, std = 0.1650), indicating less stable representations early in the network. Notably, the last layer shows a substantial drop in similarity (mean = 0.3745) and a sharp increase in variability (std = 0.3988), suggesting a divergence in representational behavior, potentially due to task-specific tuning or greater representational fragmentation. This may help explain the weaker correlations between cosine similarity and task performance observed in Qwen’s final layers.

These findings reinforce the role of middle layers in capturing semantically meaningful and transferable representations, particularly in instruction-tuned or general-purpose multilingual models. See Figure~\ref{fig:labse_qwen_histogram} for the histogram plot and Figure~\ref{fig:finding_1} for the bar chart per alignment dataset.

\subsection{Additional Analysis For Alignment and Downstream Correlation}
\label{app:additional-alignment-downstream-correlation}

As shown in Table~\ref{tab:cosine_performance_correlation}, the correlation between cosine similarity and downstream performance varies by dataset, layer, and model architecture. The following sections provide detailed interpretations.

\begin{table}[htbp]
\centering
\resizebox{\linewidth}{!}{
\begin{tabular}{lcccccc}
\toprule
\textbf{Dataset} & \textbf{Model} & \textbf{Layer} & \textbf{Pearson $r$} & \textbf{$R^2$} & \textbf{$p$-value} \\
\midrule
\multirow{6}{*}{SIB200} 
  & \multirow{3}{*}{LaBSE} 
  & First  &  0.323 & 0.104 & $<\!10^{-300}$ \\
  &        & Middle &  0.309 & 0.096 & $<\!10^{-300}$ \\
  &        & Last   &  0.210 & 0.044 & $<\!10^{-205}$ \\
  \cmidrule(lr){2-6}
  & \multirow{3}{*}{Qwen2.5-0.5B} 
  & First  &  0.060 & 0.004 & $<\!10^{-17}$ \\
  &        & Middle &  0.123 & 0.015 & $<\!10^{-69}$ \\
  &        & Last   &  0.043 & 0.002 & $<\!10^{-9}$ \\
\midrule
\multirow{6}{*}{INCLUDE-BASE} 
  & \multirow{3}{*}{LaBSE} 
  & First  & -0.041 & 0.002 & 0.233 \\
  &        & Middle &  0.005 & 0.000 & 0.884 \\
  &        & Last   & -0.021 & 0.000 & 0.545 \\
  \cmidrule(lr){2-6}
  & \multirow{3}{*}{Qwen2.5-0.5B} 
  & First  &  0.183 & 0.034 & $<\!10^{-7}$ \\
  &        & Middle &  0.142 & 0.020 & $<\!10^{-4}$ \\
  &        & Last   &  0.168 & 0.028 & $<\!10^{-6}$ \\
\midrule
\multirow{6}{*}{XCOPA} 
  & \multirow{3}{*}{LaBSE} 
  & First  & -0.115 & 0.013 & 0.458 \\
  &        & Middle & -0.026 & 0.001 & 0.867 \\
  &        & Last   &  0.144 & 0.021 & 0.352 \\
  \cmidrule(lr){2-6}
  & \multirow{3}{*}{Qwen2.5-0.5B} 
  & First  &  0.292 & 0.085 & 0.055 \\
  &        & Middle & -0.139 & 0.019 & 0.368 \\
  &        & Last   &  0.538 & 0.289 & $<\!0.001$ \\
\midrule
\multirow{6}{*}{PAWS-X} 
  & \multirow{3}{*}{LaBSE} 
  & First  &  0.141 & 0.020 & 0.484 \\
  &        & Middle &  0.270 & 0.073 & 0.173 \\
  &        & Last   &  0.146 & 0.021 & 0.467 \\
  \cmidrule(lr){2-6}
  & \multirow{3}{*}{Qwen2.5-0.5B} 
  & First  &  0.228 & 0.052 & 0.252 \\
  &        & Middle &  0.532 & 0.283 & 0.004 \\
  &        & Last   &  0.369 & 0.136 & 0.059 \\
\midrule

\end{tabular}
}
\caption{Pearson correlation coefficients ($r$), $R^2$, and $p$-values for the relationship between cosine similarity and task performance across different transformer layers on LaBSE and Qwen2.5-0.5B. 
}
\label{tab:cosine_performance_correlation}
\end{table}

\paragraph{SIB200}
For LaBSE, correlation values are consistently strong and statistically significant across all layers. The first (Pearson $r = 0.323$), middle (Pearson $r = 0.309$), and last (Pearson $r = 0.210$) layers all demonstrate meaningful positive correlations with performance $(p \approx 0)$, indicating that cosine similarity is well-aligned with task accuracy throughout the network. This suggests that SIB200 benefits from LaBSE’s cross-lingual representations, especially in the earlier and middle layers. In contrast, Qwen2.5-0.5B shows very weak but statistically significant correlations ($r \leq 0.12$ across all layers). While the trends are consistent, the effect sizes are negligible, suggesting that cosine similarity has limited practical influence on performance for Qwen2.5-0.5B on this dataset.

\paragraph{INCLUDE-BASE}
For LaBSE, correlations between cosine similarity and performance are negligible and statistically non-significant across all layers, with Pearson $r$ values close to zero ($-0.041$, $0.005$, $-0.021$). This suggests no meaningful alignment between representational similarity and task accuracy. In contrast, Qwen2.5-0.5B exhibits weak but statistically significant positive correlations (Pearson $r$ range: $0.14$–$0.18$), indicating that higher cosine similarity is marginally associated with improved performance. Despite the small effect sizes, these results highlight a slight but consistent behavioural alignment in Qwen2.5-0.5B on this dataset.

\paragraph{XCOPA}
For LaBSE, correlation values across layers are weak and statistically insignificant, suggesting minimal alignment between representational similarity and model performance. In contrast, Qwen2.5-0.5B exhibits a strong and statistically significant positive correlation in the last layer (Pearson $r = 0.538$, $p$ < 0.001), implying that deeper representations may be more predictive for XCOPA.

\paragraph{PAWS-X}
LaBSE shows weak, non-significant positive correlations across layers. However, Qwen2.5-0.5B demonstrates a strong positive correlation in the middle layer (Pearson $r = 0.532$, $p \approx 0.004$), suggesting that intermediate representations capture more alignment-relevant features for paraphrase detection.

\begin{figure*}[htbp]
    \centering
    \begin{subfigure}[b]{0.48\textwidth}
        \includegraphics[width=\textwidth]{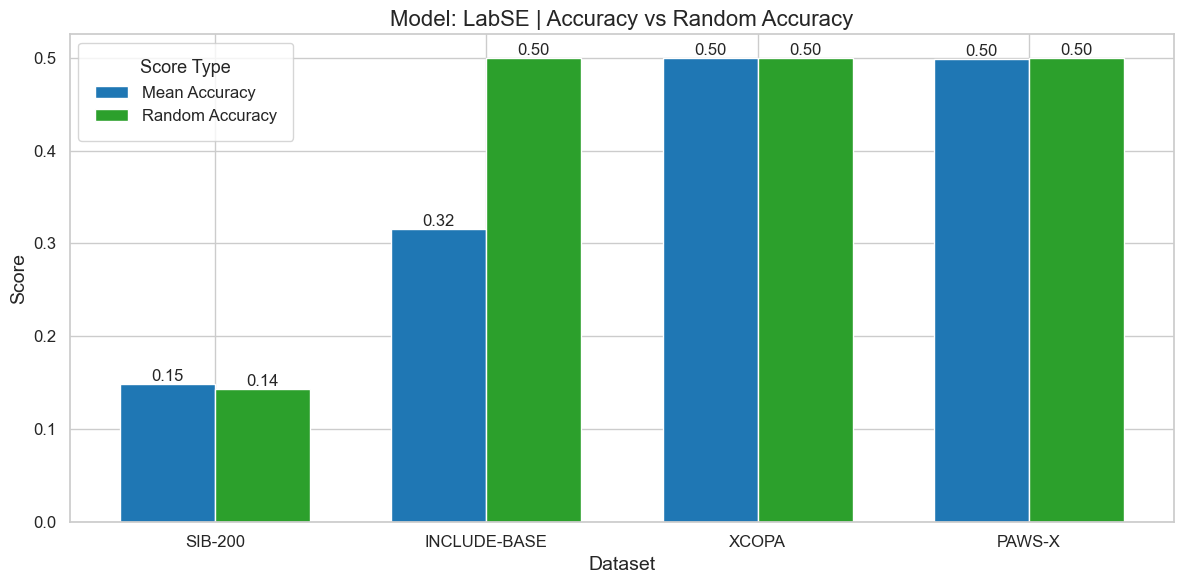}
        \caption{Performance of LaBSE across downstream tasks compared to random baselines.}
        \label{fig:finding_2_labse_bar_chart}
    \end{subfigure}
    \hfill
    \begin{subfigure}[b]{0.48\textwidth}
        \includegraphics[width=\textwidth]{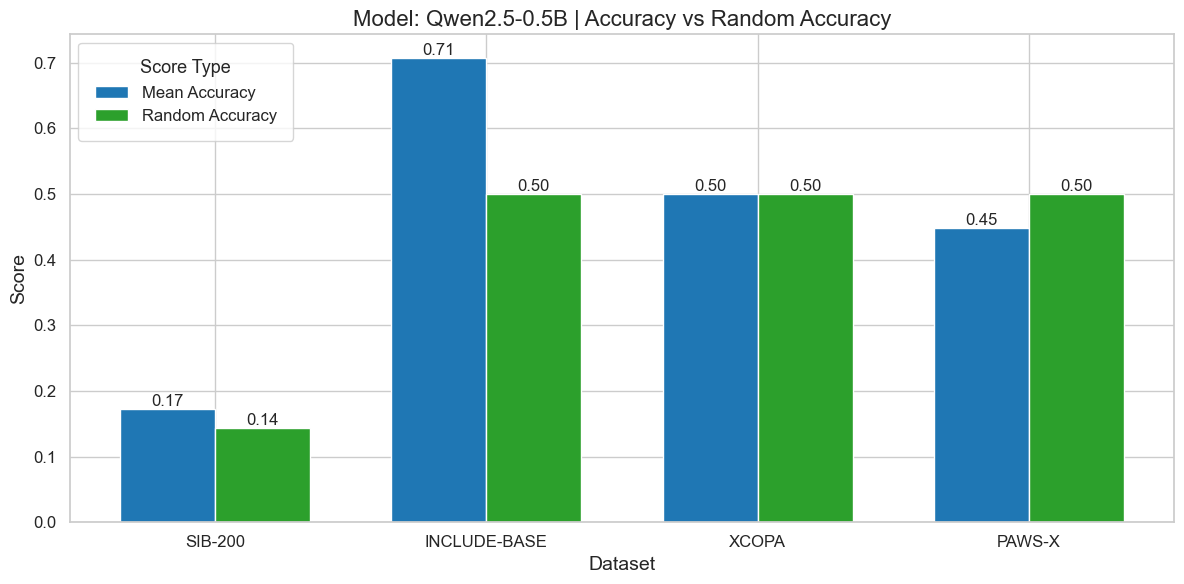}
        \caption{Performance of Qwen2.5-0.5B across downstream tasks compared to random baselines.}
        \label{fig:finding_2_qwen_bar_chart}
    \end{subfigure}
    \caption{Comparison of LaBSE and Qwen2.5-0.5B performance across various downstream tasks and their corresponding random baselines.}
    \label{fig:finding_2_labse_qwen_bar_chart}
\end{figure*}

\paragraph{Downstream Performance Relative to Random Baselines}

To provide a clearer picture of cross-lingual generalization and behavior alignment, we present a set of bar charts comparing the performance of LaBSE and Qwen2.5-0.5B across four downstream evaluation datasets—SIB200, INCLUDE-BASE, XCOPA, and PAWS-X—relative to their respective random baselines.

On XCOPA and PAWS-X, LaBSE yields near-random or below-random performance, indicating that its fixed representations struggle with cross-lingual commonsense reasoning and paraphrase detection. For SIB200, LaBSE performs slightly above the random baseline, suggesting limited task sensitivity in multilingual sentence similarity settings. However, its performance on INCLUDE-BASE remains weak, staying near or below the random baseline and highlighting deficiencies in broader multilingual alignment.

In contrast, Qwen2.5-0.5B demonstrates stronger generalization on both SIB200 and INCLUDE-BASE, significantly outperforming its baseline and showing evidence of better cross-lingual task adaptation. However, it faces challenges on XCOPA and PAWS-X, where its performance hovers around or falls below baseline, pointing to possible limitations in zero-shot commonsense reasoning and paraphrase understanding across languages.

These comparisons highlight the differing strengths and weaknesses of encoder-only and decoder-only multilingual models across select zero-shot evaluation tasks. See Figure~\ref{fig:finding_2_labse_qwen_bar_chart}.

\subsection{Additional Analysis For Alignment and LID Correlation}

As shown in Table~\ref{tab:lid_alignment_correlation}, the correlation between alignment (as measured by cosine similarity) and downstream LID performance varies notably across datasets, model architectures, and transformer layers. The following sections provide detailed interpretations for each dataset to contextualize these trends.

\begin{table}[!t]
\centering
\small
\resizebox{\linewidth}{!}{
\begin{tabular}{lcccccc}
\toprule
\textbf{Dataset} & \textbf{Model} & \textbf{Layer} & \textbf{Pearson $r$} & \textbf{$R^2$} & \textbf{$p$-value} \\
\midrule
\multirow{6}{*}{FLORES-200} 
  & \multirow{3}{*}{LaBSE} 
  & First  &  0.024 & 0.001 & 0.732\\
  &        & Middle &  -0.122 & 0.015 & 0.084\\
  &        & Last   &  -0.707 & 0.500 & $<\!10^{-31}$ \\
  \cmidrule(lr){2-6}
  & \multirow{3}{*}{Qwen2.5-0.5B} 
  & First  &  -0.142 & 0.020 & 0.043\\
  &        & Middle &  -0.432 & 0.186 & $<\!10^{-9}$ \\
  &        & Last   &  -0.278 & 0.077 & $<\!10^{-4}$ \\
\midrule
\multirow{6}{*}{NTREX-128} 
  & \multirow{3}{*}{LaBSE} 
  & First  & 0.254 & 0.065 & 0.012\\
  &        & Middle & -0.173 & 0.030 & 0.089 \\
  &        & Last   & -0.621 & 0.385 & $<\!10^{-11}$\\
  \cmidrule(lr){2-6}
  & \multirow{3}{*}{Qwen2.5-0.5B} 
  & First  &  -0.232 & 0.054 & 0.021 \\
  &        & Middle &  -0.476 & 0.226 & $<\!10^{-6}$ \\
  &        & Last   &  -0.340 & 0.115 & 0.001\\
\midrule
\multirow{6}{*}{NusaX} 
  & \multirow{3}{*}{LaBSE} 
  & First  & -0.566 & 0.320 & 0.112 \\
  &        & Middle &  -0.872 & 0.760 & 0.002 \\
  &        & Last & -- & -- & -- \\
  \cmidrule(lr){2-6}
  & \multirow{3}{*}{Qwen2.5-0.5B} 
  & First  &  -0.455 & 0.207 & 0.218 \\
  &        & Middle &  -0.873 & 0.763 & 0.002 \\
  &        & Last   &  -0.045 & 0.002 & 0.910\\
\bottomrule
\end{tabular}
}
\caption{Pearson correlation coefficients ($r$), $R^2$, and $p$-values for the relationship between KNN LID F1 score using mean-pooled embedding and alignment cosine similarity across different transformer layers on LaBSE and Qwen2.5-0.5B.}
\label{tab:lid_alignment_correlation}
\end{table}

\paragraph{FLORES-200}
On the FLORES-200 dataset, we observe a moderate negative correlation between cosine similarity and LID performance for both LaBSE and Qwen2.5-0.5B. The strength of the correlation increases in deeper layers, with the last layer showing the strongest correlation ($r = -0.707$, $p < 10^{-31}$) for LaBSE. Qwen2.5-0.5B, however, exhibits its strongest negative correlation in the middle layer ($r = -0.432$, $p < 10^{-9}$), indicating that as the embeddings become more aligned (i.e., higher cosine similarity), the language identity signal tends to weaken, potentially due to semantic abstraction. The statistically significant $p$-values across all layers confirm the robustness of this relationship. These findings reinforce the idea that high alignment may come at the cost of LID separability, especially in final layers for LaBSE and middle layer for Qwen2.5-0.5B, where representations are more semantically homogenized.

\paragraph{NTREX-128}
For NTREX-128, the correlation trends diverge between the two models. LaBSE exhibits its strongest negative correlation in the the last layer (Pearson $r = -0.621$, $p < 10^{-11}$), with a positive correlation in the first layer (Pearson $r = 0.254$, $p = 0.012$) and weak negative correlation in the middle (Pearson $r = -0.173$, $p = 0.089$). This suggests that early representations in LaBSE may still retain relatively distinct language features that diminish with depth. 
In contrast, Qwen2.5-0.5B shows more consistent negative correlations across all layers, particularly in the middle layer (Pearson $r = -0.476$, $p < 10^{-6}$). These results highlight a more uniform degradation of LID-relevant information in Qwen’s architecture compared to LaBSE.

\paragraph{NusaX}
For NusaX, alignment-LID correlations exhibit distinct patterns. LaBSE shows a weak correlation in the first layer (Pearson $r = -0.566$, $p = 0.112$), a highly negative correlation in the middle layer (Pearson $r = -0.872$, $p = 0.002$), and no measurable correlation in the last layer (--), which we assume reflects a perfect inverse relationship (Pearson $r \approx -1$) due to complete LID failure. Qwen2.5-0.5B follows a similar pattern, with its most negative correlation in the middle layer (Pearson $r = -0.873$, $p = 0.002$) and negligible correlations in the first (Pearson $r = -0.455$, $p = 0.218$) and last layers (Pearson $r = -0.045$, $p = 0.910$). The correlations for both models are the most negative observed across all datasets, suggesting alignment disproportionately degrades language signals in low-resource settings. This extreme inverse relationship likely stems from the models’ lack of prior exposure to NusaX languages during training, limiting their ability to retain language identity in aligned embeddings.

\section{LID Methods and Results}
\label{app:lid_appendix}
\begin{table*}[ht]
  \centering
  \small
  \label{tab:f1-lang-id-all}
  \begin{tabular}{lll *{3}{cc}}
    \toprule
    & & & \multicolumn{2}{c}{FLORES-200}
          & \multicolumn{2}{c}{NTREX-128}
          & \multicolumn{2}{c}{NusaX} \\
    \cmidrule(lr){4-5} \cmidrule(lr){6-7} \cmidrule(lr){8-9}
    \textbf{Model} & \textbf{Method} & \textbf{Layer}
      & \textbf{CLS} & \textbf{Mean}
      & \textbf{CLS} & \textbf{Mean}
      & \textbf{CLS} & \textbf{Mean} \\
    \midrule
    \multirow{6}{*}{LaBSE}
      & \multirow{3}{*}{KNN}
                                 & First  & 80.65 & 88.35 & 87.02 & 90.43 & 64.12 & 81.78 \\
      &                          & Middle & 65.11 & 78.85 & 71.37 & 81.30 & 33.89 & 45.37 \\
      &                          & Last   & 7.65  & 3.92  & 3.45  & 1.63  & 0.54  & 0.00 \\
      \cmidrule(lr){2-9}
      & \multirow{3}{*}{\parbox{4cm}{Linear\\Probing}}
                                 & First  & 93.47 & 95.13 & 92.21 & 93.29 & 89.16 & 97.30 \\
      &                          & Middle & 92.99 & 94.18 & 92.33 & 92.68 & 88.00 & 94.51 \\
      &                          & Last   & 30.03 & 70.89 & 22.91 & 74.36 & 56.00 & 65.44 \\
    \midrule
    \multirow{6}{*}{Qwen2.5-0.5B}
      & \multirow{3}{*}{KNN}
                                 & First  & –    & 83.69 & –    & 86.06 & –    & 65.79 \\
      &                          & Middle & –    & 55.32 & –    & 54.73 & –    & 25.05 \\
      &                          & Last   & –    & 71.73 & –    & 81.86 & –    & 29.39 \\
      \cmidrule(lr){2-9}
      & \multirow{3}{*}{\parbox{4cm}{Linear\\Probing}}
                                 & First  & –    & 94.21 & –    & 91.42 & –    & 95.55 \\
      &                          & Middle & –    & 91.76 & –    & 90.04 & –    & 87.09 \\
      &                          & Last   & –    & 92.46 & –    & 90.27 & –    & 88.77 \\
    \bottomrule
  \end{tabular}
\caption{F1 score for KNN and linear classifiers by layer and pooling on FLORES-200, NTREX-128, and NusaX.}
\label{tab:lid-fl-full}
\end{table*}

\subsection{Methods}
To investigate language-specific information in multilingual representations, we analyze two distinct paradigms: (1) frozen embeddings from pretrained decoder-only LLMs (Qwen-2.5) and (2) specialized multilingual sentence encoders (LaBSE). We evaluate whether linguistic identity is recoverable from their hidden states and how pooling strategies affect clusterability (via non-parametric KNN retrieval) and linear separability (via supervised classification heads).
\paragraph{KNN-based Language Identification}
We hypothesize that language identity manifests as separable clusters in the hidden space, which can be detected via non-parametric nearest-neighbor retrieval.

For both Qwen-2.5 and LaBSE, hidden states are extracted from the first ($\ell=1$), middle ($\ell=m$), and final ($\ell=L$) layers. Let \(\mathbf{H}^\ell \in \mathbb{R}^{T \times d}\) denote the hidden states at layer \(\ell\) for a sequence of length \(T\). Sentence-level embeddings are derived as follows:
\begin{itemize}
    \item Qwen-2.5: Only mean pooling is applied:
    \[
    \mathbf{e}^{\ell}_{\text{mean}} = \frac{1}{T} \sum_{t=1}^{T} \mathbf{H}_t^{\ell} \in \mathbb{R}^d.
    \]
    \item LaBSE: Both CLS and mean pooling are compared:
    \[
    \mathbf{e}^{\ell}_{\text{CLS}} = \mathbf{H}^{\ell}_{[CLS]}, \quad 
    \mathbf{e}^{\ell}_{\text{mean}} = \frac{1}{T} \sum_{t=1}^{T} \mathbf{H}_t^{\ell} \in \mathbb{R}^d.
    \]
\end{itemize}
For each layer \(\ell \in \{1, m, L\}\) and pooling strategy \(\mathrm{pool} \in \{\text{mean}, \text{CLS}\}\), we construct reference sets:
\[
\mathcal{R}^\ell_{\mathrm{pool}} = \left\{ \left( \mathbf{e}^{\ell,(i,j)}_{\mathrm{pool}}, y^{(j)} \right) \right\}_{i=1,j=1}^{200, 204},
\]
where \(y^{(j)}\) is the language label for the \(j\)-th language in FLORES-200, and \(i\) indexes the examples within each language. This results in a total of \(200 \times 204 = 40,800\) reference embeddings. For Qwen-2.5, only \(\mathcal{R}^\ell_{\text{mean}}\) is used, while LaBSE employs both \(\mathcal{R}^\ell_{\text{CLS}}\) and \(\mathcal{R}^\ell_{\text{mean}}\).

We evaluate on three test sets: Flores-200, NTREX-128, and NusaX. To ensure fair comparison, we retain only languages overlapping with the FLORES-200 train set:
\[
\mathcal{L}_{\text{overlap}} = \mathcal{L}_{\text{test}} \cap \mathcal{L}_{\text{FLORES-train}},
\]
where \(\mathcal{L}_{\text{test}}\) is the language set of the test dataset, and \(\mathcal{L}_{\text{FLORES-train}}\) contains the 204 languages in the FLORES-200 train set. For a test embedding \(\mathbf{e}^{\ell}_{\text{test,pool}}\), we compute its L2 distance to all reference embeddings in \(\mathcal{R}^\ell_{\mathrm{pool}}\):
\[
\begin{split}
d\left(\mathbf{e}^{\ell}_{\text{test,pool}}, \mathbf{e}^{\ell,(i,j)}_{\text{ref,pool}}\right) &= \left\| \mathbf{e}^{\ell}_{\text{test,pool}} - \mathbf{e}^{\ell,(i,j)}_{\text{ref,pool}} \right\|_2^2, \\
&\quad \forall i \in \{1, \ldots, 200\}, \\
&\quad \forall j \in \{1, \ldots, 204\}.
\end{split}
\]

The predicted language \(\hat{y}_{\text{test}}\) is obtained via majority vote over the \(k=256\) nearest neighbors:
\[
\hat{y}_{\text{test}} = \underset{l \in \mathcal{L}_{\text{overlap}}}{\arg\max} \sum_{(i,j) \in \mathcal{N}_k} \mathbf{1}(y^{(j)} = l),
\]
where \(\mathcal{N}_k\) denotes the set of indices for the top-\(k\) neighbors, and \(\mathbf{1}\) is the indicator function.

\paragraph{Linear Classification Head} To complement our non-parametric analysis, we probe the linear separability of language identity in Qwen-2.5 and LaBSE embeddings. This evaluates whether linguistic boundaries are geometrically aligned with hyperplanes in the hidden space, which would suggest that language control can be achieved through simple affine transformations.

Similar to the KNN-based approach, embeddings are extracted identically. For each dataset \(\mathcal{D} \in \{\text{FLORES-200, NTREX-128, NusaX}\}\) and each layer \(\ell \in \{1, m, L\}\) representing early, middle, and last layers respectively, we train a separate linear layer to map embeddings \(\mathbf{e}^{\ell} \in \mathbb{R}^d\) to language logits \(\mathbf{z}^{\ell} \in \mathbb{R}^C\), where \(C\) is the number of languages. The classifier for each layer is defined as:
\[
\mathbf{z}^{\ell} = \mathbf{W}^{\ell}\mathbf{e}^{\ell} + \mathbf{b}^{\ell}, \quad \mathbf{W}^{\ell} \in \mathbb{R}^{C \times d}, \mathbf{b}^{\ell} \in \mathbb{R}^C,
\]
with cross-entropy loss minimized during training.

\subsection{Results}

Our analysis reveals distinct layer-wise behaviors in language identification (LID) performance across LaBSE and Qwen2.5-0.5B models, focus on mean-pooled embedding. 

\paragraph{KNN-based Language Identification}
The KNN method highlights significant performance variations across layers. As shown in Table~\ref{tab:mean-pooled-lid-fl}, for LaBSE, the first layer achieves robust results, with mean F1 scores of 88.35\% on FLORES-200, 90.43\% on NTREX-128, and 81.78\% on NusaX. Performance declines moderately in the middle layer, yielding 78.85\% for FLORES-200, 81.30\% for NTREX-128, and 45.37\% for NusaX. The last layer exhibits catastrophic degradation, collapsing to 3.92\%, 1.63\%, and 0.00\% on the respective datasets. This suggests that deeper LaBSE layers lose language-discriminative features critical for KNN classification.

For Qwen2.5-0.5B, the first layer similarly outperforms middle layers, with mean F1 scores of 83.69\% on FLORES-200, 86.06\% on NTREX-128, and 65.79\% on NusaX. The middle layer shows the weakest results across all datasets: 55.32\%, 54.73\%, and 25.05\%, respectively, while the last layer partially recovers to 71.73\%, 81.86\%, and 29.39\%. This non-monotonic trend suggests limited retention of language-specific signals in the middle layer of Qwen2.5-0.5B.

LaBSE, trained for semantic alignment, shows severe degradation in its final layer, with near-zero F1 scores across datasets, as deeper layers erase language-specific signals required for KNN classification. In contrast, Qwen2.5-0.5B, a standard pretrained LLM, experiences a performance dip in its middle layer but recovers partially in the final layer, retaining sufficient linguistic discriminability. This divergence underscores a key architectural trade-off: contrastive models like LaBSE discard lexical or syntactic patterns in deeper layers to prioritize semantic invariance, while standard LLMs preserve partial language-identifying features across layers despite progressive abstraction.

\paragraph{Linear-probing-based Language Identification}
For LaBSE, the First Layer consistently achieves the highest LID F1 scores across all datasets, with a significant drop in performance observed in the Last Layer. The NusaX dataset delivers the best overall results, particularly in the First Layer, where it reaches 97.30\% F1 score. However, the Last Layer shows notably weaker performance, especially for the FLORES-200 and NusaX datasets. These findings suggest that earlier layers of LaBSE retain more language-identification-relevant features, such as surface-level linguistic cues, compared to deeper layers (see Table~\ref{tab:mean-pooled-lid-fl}).

Similarly, in the Qwen2.5-0.5B model, the First Layer consistently outperforms the Middle Layer in LID F1 scores across all datasets. The NusaX dataset again produces the best results, with 95.55\% F1 score, while NTREX-128 exhibits the lowest performance across all layers. These results indicate that the shallow First Layer of Qwen2.5-0.5B is more effective for language identification tasks than deeper layers, such as the Middle Layer, which shows weaker performance (refer to Table~\ref{tab:mean-pooled-lid-fl}).

Overall, both models show that their highest LID performance occurs in the First Layer, with F1 scores declining as the layers get deeper. The NusaX dataset consistently yields the best performance, while the Last Layer in LaBSE and the Middle Layer in Qwen2.5-0.5B exhibit the weakest results. These trends suggest that shallow layers retain more language-specific information, which is crucial for language identification, likely due to their greater focus on surface-level features and general linguistic patterns. Table~\ref{tab:lid-fl-full} further illustrate the comparative performance across layers and pooling techniques for both LaBSE and Qwen2.5-0.5B models.

\paragraph{Classifier Comparison: KNN vs. Linear Head}
As shown in Table~\ref{tab:lid-fl-full}, linear classifiers achieve superior F1 scores compared to KNN across layers, suggesting their ability to identify language-discriminative features within linearly separable subspaces. However, linear methods exhibit attenuated performance gaps between layers, for instance, the difference between first and middle layers in Qwen2.5-0.5B is less than 5\% with linear classifiers, while KNN reveals differences exceeding 30\%. Similarly, LaBSE’s linear classifier reduces the last-layer performance gap to under 25\%, whereas KNN shows near-complete degradation. This contrast implies that parametric linear methods, while more accurate overall, may obscure layer-specific language information degradation due to their reliance on learned projections. In contrast, KNN’s non-parametric nature might more directly reflect the geometric structure of embeddings, amplifying sensitivity to layer-wise shifts in language information quality.

\paragraph{Pooling Method Comparison: CLS Token vs. Mean}  
As shown in Table~\ref{tab:lid-fl-full}, the effectiveness of pooling strategies varies across layers. In first and middle layers, mean pooling achieves superior performance, with F1 margins exceeding 10\% over CLS token pooling under KNN. However, in last layers, CLS token pooling shows limited resilience under KNN, marginally outperforming mean pooling in isolated cases despite near-random overall performance. Linear classifiers amplify mean pooling’s advantage across all layers, suggesting its robustness to layer-specific degradation.  

This suggests that mean pooling better preserves language-discriminative signals across layers, likely due to its aggregation of token-level features. In contrast, the CLS token, optimized for semantic tasks, exhibits sharper performance declines in deeper layers, particularly under non-parametric methods like KNN. These observations highlight the interplay between pooling strategy, layer depth, and classification method in language identification tasks. 

\section{Language Vector Setting}
\label{app:lda-in-details}
\begin{figure*}    
    \centering
    \begin{minipage}{0.37\linewidth}
        \resizebox{\linewidth}{!}{
            \includegraphics[trim=0 0 0 0,clip]{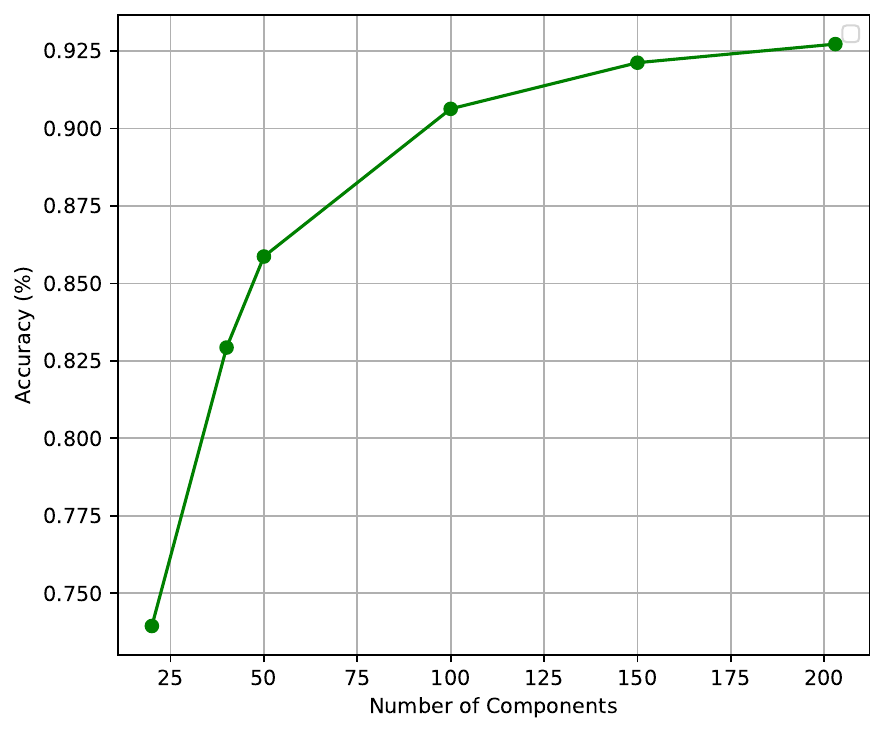}
        }
    \end{minipage}
    \hspace{6pt}
    \begin{minipage}{0.37\linewidth}
        \resizebox{\linewidth}{!}{
            \includegraphics[trim=0 0 0 0,clip]{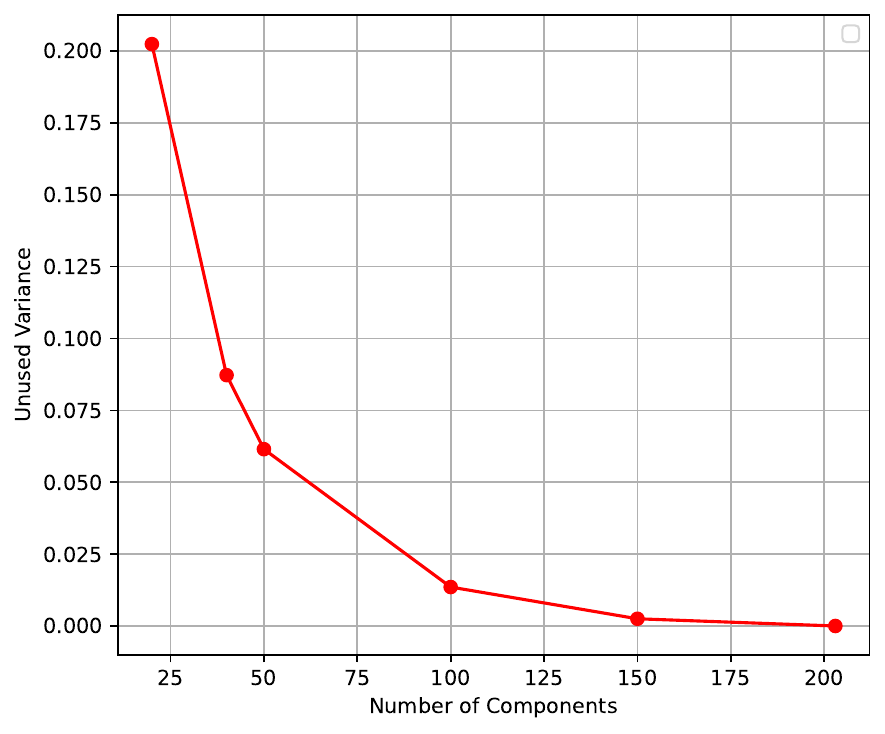}
        }
    \end{minipage}
    \caption{Controlling the number of language feature representations by using LDA performance accuracy (\textbf{Left}) and unused variance (\textbf{Right}) across number of components.}
    \label{fig:lda_acc_variance}
\end{figure*}

Linear Discriminant Analysis (LDA)~\cite{balakrishnama1998lda,tharwat2017lda} is utilized to construct language vectors by extracting language-specific features from the Qwen2.5-0.5B model’s scaled hidden states, optimizing cross-lingual control through class separability. We evaluate various component sizes (20, 40, 50, 100, 150, 203) to balance LID accuracy and unused variance, fitting an LDA model and training a linear neural network (with 10 epochs, Adam optimizer, and CrossEntropyLoss) to achieve a peak accuracy of approximately 90.63\% at 100 components. The unused variance is minimized, ensuring retained discriminative information for injection (\(\mathbf{\delta}\)) with pruning, which enhances language targeting while the Figure \ref{fig:lda_acc_variance} visually confirms this optimal trade-off.

\section{Ablation on Language Shift Strategy}
\label{app:lang-shift-strategy}

\paragraph{Language Shift Strategy}

We assess various strategies for injecting the language vector in \methodname{}. Specifically, we explore three strategies based on the temporal scope of the latent intervention: (1) prompt only, (2) generated tokens only, and (3) both phases. Let \(\mathbf{h}_t^{(m)} \in \mathbb{R}^d\) denote the hidden state at position \(t\) in the middle layer \(m\), and \(\mathbf{h}_t^{(m)'}\) denotes its language-shifted counterpart:

\begin{itemize}
    \item \textbf{Prompt-Only} (\texttt{prompt-only}): Applies injection exclusively to input prompt processing:
    \[
    \mathbf{h}_t^{(m)'} = \begin{cases}
        \mathbf{h}_t^{(m)} + \alpha\mathbf{\delta}, & \forall t \in [1, T_{\text{input}}] \\
        \mathbf{h}_t^{(m)}, & \forall t > T_{\text{input}}
    \end{cases}
    \]

    \item \textbf{Generated-Only} (\texttt{gen-only}): Restricts injection to autoregressive generation:
    \[
    \resizebox{\linewidth}{!}{$
    \mathbf{h}_t^{(m)'} = \begin{cases}
        \mathbf{h}_t^{(m)}, & \forall t \in [1, T_{\text{input}}] \\
        \mathbf{h}_t^{(m)} + \alpha\mathbf{\delta}, & \forall t \in [T_{\text{input}}+1, T_{\text{total}}]
    \end{cases}
    $}
    \]

    \item \textbf{Prompt and Generated} (\texttt{prompt-and-gen}): Applies injection throughout both phases:
    \[
    \mathbf{h}_t^{(m)'} = \mathbf{h}_t^{(m)} + \alpha\mathbf{\delta}, \quad \forall t \in [1, T_{\text{total}}]
    \]
    
\end{itemize}

where \(T_{\text{input}}\) is the input prompt length and \(T_{\text{total}} = T_{\text{input}} + N\) the total sequence length after generating \(N\) tokens.

\paragraph{Ablation Result}

\begin{figure}[!t]
\centering
\includegraphics[width=\linewidth]{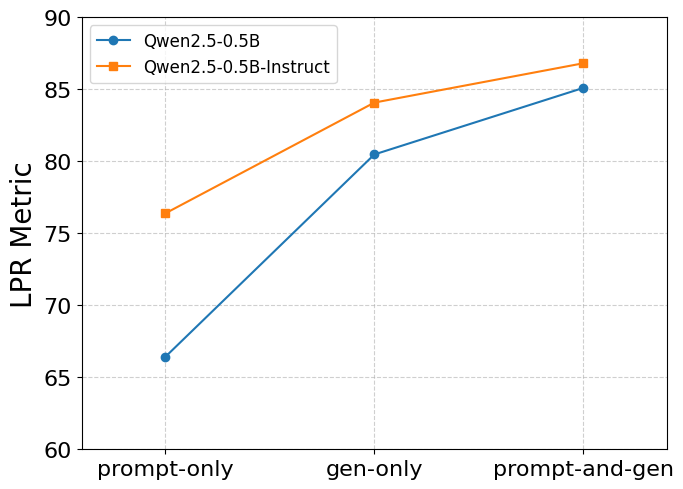}
\caption{Cross-lingual LPR performance across different vector injection strategies.}
\label{fig:lpr-vector-timestep}
\end{figure}

All three language shift strategies are compared in cross-lingual setting using the Qwen2.5-0.5B and Qwen2.5-0.5B-Instruct, as shown in Figure~\ref{fig:lpr-vector-timestep}. The \textbf{prompt-and-gen} strategy consistently achieves the strongest performance, followed by \textbf{gen-only} and then \textbf{prompt-only}. This indicates that while the \textbf{prompt-only} approach may aid the model in understanding the input context in the target language, and the \textbf{gen-only} strategy directly shifts the generation process into target language, while the \textbf{prompt-and-gen} method effectively combines both advantages via injecting the shift language vector into all timesteps.

\begin{table*}[!t]
\centering
\resizebox{\linewidth}{!}{
    \begin{tabular}{c|cc|cc|cc|cc|cc|cc}
    \toprule
    \multirow{3}{*}{\textbf{Lang}} &
    \multicolumn{4}{c|}{\textbf{Qwen2.5-0.5B}} &
    \multicolumn{4}{c|}{\textbf{Qwen2.5-7B}} &
    \multicolumn{4}{c}{\textbf{Llama-3.1-8B}} \\
    \cmidrule(lr){2-5} \cmidrule(lr){6-9} \cmidrule(lr){10-13}
    & \multicolumn{2}{c|}{Baseline} & \multicolumn{2}{c|}{\methodname{}} & 
      \multicolumn{2}{c|}{Baseline} & \multicolumn{2}{c|}{\methodname{}} & 
      \multicolumn{2}{c|}{Baseline} & \multicolumn{2}{c}{\methodname{}} \\
    \cmidrule(lr){2-3} \cmidrule(lr){4-5} \cmidrule(lr){6-7} \cmidrule(lr){8-9} \cmidrule(lr){10-11} \cmidrule(lr){12-13}
    & chrF++ & BERT F1 & chrF++ & BERT F1 & chrF++ & BERT F1 & chrF++ & BERT F1 & chrF++ & BERT F1 & chrF++ & BERT F1 \\
    \midrule
    ID & 7.71 & 61.38 & \textbf{8.46} & \textbf{63.74} & 8.21 & 62.98 & \textbf{9.26} & \textbf{65.19} & 8.63 & 60.58 & \textbf{8.91} & \textbf{64.94} \\
    TH & 3.39 & 62.12 & \textbf{3.42} & \textbf{63.78} & 3.62 & 62.55 & \textbf{3.88} & \textbf{63.90} & 2.96 & 59.02 & \textbf{4.28} & \textbf{64.37} \\
    TR & 6.42 & 59.36 & \textbf{6.78} & \textbf{60.67} & 6.94 & 61.31 & \textbf{7.59} & \textbf{62.9}6 & 8.37 & 58.20 & \textbf{8.62} & \textbf{63.76} \\
    JA & 1.90 & 59.98 & \textbf{2.11} & \textbf{61.53} & \textbf{2.08} & 60.14 & 1.84 & \textbf{61.15} & 1.52 & 53.26 & \textbf{2.60} & \textbf{62.94} \\
    FR & 7.53 & 61.63 & \textbf{8.89} & \textbf{64.03} & 8.11 & 63.03 & \textbf{9.51} & \textbf{65.24} & 7.97 & 59.86 & \textbf{8.90} & \textbf{64.51} \\
    ES & 8.51 & 62.66 & \textbf{9.43} & \textbf{64.90} & 9.30 & 64.24 & \textbf{10.01} & \textbf{65.65} & 8.69 & 61.14 & \textbf{9.84} & \textbf{65.65} \\
    AR & 5.11 & 61.89 & \textbf{5.68} & \textbf{64.31} & 5.35 & 62.39 & \textbf{6.78} & \textbf{65.98} & 4.28 & 59.70 & \textbf{6.45} & \textbf{65.59} \\
    KO & 1.86 & 60.93 & \textbf{2.08} & \textbf{61.90} & 2.09 & 61.67 & \textbf{2.14} & \textbf{62.35} & 2.14 & 54.61 & \textbf{3.31} & \textbf{65.19} \\
    ZH & 2.61 & 62.26 & \textbf{2.97} & \textbf{64.85} & 2.73 & 62.93 & \textbf{3.33} & \textbf{65.18} & 2.00 & 55.14 & \textbf{2.67} & \textbf{63.98} \\
    \midrule
    AVG & 5.01 & 61.36 & \textbf{5.53} & \textbf{63.30} & 5.38 & 62.36 & \textbf{6.04} & \textbf{64.18} & 5.39 & 58.39 & \textbf{6.76} & \textbf{64.29} \\
    \bottomrule
    \end{tabular}
}
\caption{Generation performance for different target languages on Qwen2.5 and Llama-3.1 base version. \textbf{Baseline} denotes the same model prompted in the same language as the desired target language. Bold values indicate the best score for each metric across all models and settings.}
\label{tab:cross-lingual-base}
\end{table*}

\begin{table*}[!t]
\centering
\resizebox{\linewidth}{!}{
    \begin{tabular}{c|cc|cc|cc|cc|cc|cc}
    \toprule
    \multirow{3}{*}{\textbf{Lang}} &
    \multicolumn{4}{c|}{\textbf{Qwen2.5-0.5B-Instruct}} &
    \multicolumn{4}{c|}{\textbf{Qwen2.5-7B-Instruct}} &
    \multicolumn{4}{c}{\textbf{Llama-3.1-8B-Instruct}} \\
    \cmidrule(lr){2-5} \cmidrule(lr){6-9} \cmidrule(lr){10-13}
    & \multicolumn{2}{c|}{Baseline} & \multicolumn{2}{c|}{\methodname{}} & \multicolumn{2}{c|}{Baseline} & \multicolumn{2}{c|}{\methodname{}} & \multicolumn{2}{c|}{Baseline} & \multicolumn{2}{c}{\methodname{}} \\
    \cmidrule(lr){2-3} \cmidrule(lr){4-5} \cmidrule(lr){6-7} \cmidrule(lr){8-9} \cmidrule(lr){10-11} \cmidrule(lr){12-13}
    & chrF++ & BERT F1 & chrF++ & BERT F1 & chrF++ & BERT F1 & chrF++ & BERT F1 & chrF++ & BERT F1 & chrF++ & BERT F1 \\
    \midrule
    ID & 7.71 & 61.38 & \textbf{8.46} & \textbf{63.74} & 8.21 & 62.98 & \textbf{9.26} & \textbf{65.19} & 9.67 & 64.58 & \textbf{11.55} & \textbf{66.97} \\
    TH & 3.39 & 62.12 & \textbf{3.42} & \textbf{63.78} & 3.62 & 62.55 & \textbf{3.88} & \textbf{63.90} & 5.42 & 64.68 & \textbf{6.67} & \textbf{67.82} \\
    TR & 6.42 & 59.36 & \textbf{6.78} & \textbf{60.67} & 6.94 & 61.31 & \textbf{7.59} & \textbf{62.96} & 9.37 & 63.37 & \textbf{10.49} & \textbf{65.15} \\
    JA & 1.90 & 59.98 & \textbf{2.11} & \textbf{61.53} & \textbf{2.08} & 60.14 & 1.84 & \textbf{61.15} & 3.33 & 63.29 & \textbf{4.11} & \textbf{66.23} \\
    FR & 7.53 & 61.63 & \textbf{8.89} & \textbf{64.03} & 8.11 & 63.03 & \textbf{9.51} & \textbf{65.24} & 9.44 & 64.28 & \textbf{11.40} & \textbf{67.52} \\
    ES & 8.51 & 62.66 & \textbf{9.43} & \textbf{64.90} & 9.30 & 64.24 & \textbf{10.01} & \textbf{65.65} & 10.32 & 64.78 & \textbf{12.24} & \textbf{67.68} \\
    AR & 5.11 & 61.89 & \textbf{5.68} & \textbf{64.31} & 5.35 & 62.39 & \textbf{6.78} & \textbf{65.98} & 6.88 & 64.82 & \textbf{8.66} & \textbf{67.55} \\
    KO & 1.86 & 60.93 & \textbf{2.08} & \textbf{61.90} & 2.09 & 61.67 & \textbf{2.14} & \textbf{62.35} & 3.74 & 64.52 & \textbf{4.59} & \textbf{66.99} \\
    ZH & 2.61 & 62.26 & \textbf{2.97} & \textbf{64.85} & 2.73 & 62.93 & \textbf{3.33} & \textbf{65.18} & 2.58 & 64.26 & \textbf{3.70} & \textbf{66.82} \\
    \midrule
    AVG & 5.41 & 61.79 & \textbf{6.11} & \textbf{63.74} & 5.82 & 63.24 & \textbf{6.48} & \textbf{64.96} & 6.97 & 64.80 & \textbf{8.49} & \textbf{67.19} \\
    \bottomrule
    \end{tabular}
}
\caption{Generation performance for different target languages on Qwen2.5 and Llama-3.1 Instruction version. \textbf{Baseline} denotes the same model prompted in the same language as the desired target language. Bold values indicate the best score within each model, and the overall best across models.}
\label{tab:cross-lingual-instruct}
\end{table*}

\begin{figure*}[h]
    \centering
    \includegraphics[width=\linewidth]{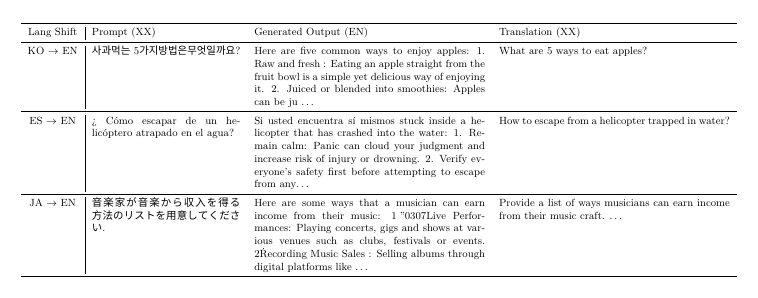} 
    \caption{Examples of generated outputs from Llama-3.1-8B-Instruct with injection in XX$\rightarrow$EN.}
    \label{fig:example_xx2en}
\end{figure*}

\begin{figure*}[h]
    \centering
    \includegraphics[width=\linewidth]{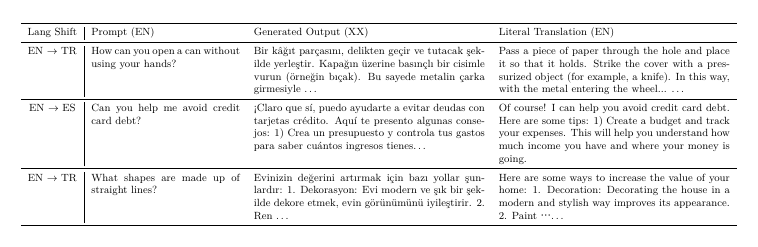} 
    \caption{Examples of generated outputs from Llama-3.1-8B-Instruct with injection in EN$\rightarrow$XX.}
    \label{fig:example_en2xx}
\end{figure*}

\section{Experiement Settings for Language Confusion}
\label{app:exp_set_lang_confusion}
\subsection{Baseline}
The results discussed is focus on Line-level Pass Rate (LPR). Word-level Pass Rate (WPR) is mostly excluded in discussion because WPR for Latin-script languages is compromised by its fundamental reliance on Unicode character ranges, a limitation highlighted in~\cite{marchisio-etal-2024-understanding}. For Latin-script WPR evaluation, we use the following Unicode ranges: Basic Latin, Latin-1 Supplement, Latin Extended-A through Latin Extended-G, and Latin Extended Additional~\footnote{Taken from Wikipedia: \url{https://en.m.wikipedia.org/wiki/Unicode_block}.}. We use the following generation hyperparameters: \texttt{max\_new\_tokens}=256 and \texttt{top\_k}=50. We apply nucleus sampling with \texttt{top\_p}=0.9 and use a moderate temperature of 0.7.

\subsection{In-context learning (ICL)}
\label{app:icl_exp_setting}
We follow all the original settings for ICL in the LCB benchmark. For the Q/A template, we use the \texttt{Q: A:} format, while the chat template adopts the model-specific instruction-tuning structure. Cross-lingual few-shot prompts follow the benchmark’s original setup, where English inputs include instructions such as \texttt{Respond in <TARGET\_LANG>}. For monolingual few-shot prompts, both inputs and outputs are in the same language.It is important to note that the few-shot demonstrations are provided in languages distinct from the target language of the current task. This design aims to guide the model in understanding the task semantics rather than identifying the appropriate output language.

\subsection{Inference-Time Language Control (\methodname{})}
\label{app:itlc_exp_setting}
We use the following scaling factor $\alpha$ values: for Qwen2.5-0.5B and Qwen2.5-0.5B-Instruct, $\alpha = 0.5$; for Qwen2.5-7B and Qwen2.5-7B-Instruct, $\alpha = 1.3$.
For Llama-3.1-8B, $\alpha = 0.15$; and for Llama-3.1-8B-Instruct, $\alpha = 0.10$. See Appendix~\ref{app:ablation_scaling} for details on the selection of scaling factor.

\subsection{Parameter-Efficient Fine-Tuning (PEFT)}
\label{app:peft_exp_setting}
For training data preparation, we use the Alpaca subset from the Bactrian-X dataset~\cite{li2023bactrianx}. Specifically, we extract the first 1,000 rows per language across the 14 languages included in the LCB benchmark (500 for monolingual and 500 for cross-lingual), resulting in a total of 14,000 samples. For monolingual data, we directly use the same samples as in Bactrian-X. For cross-lingual data, we replace the inputs with their corresponding English variants in Bactrian-X and append the instruction \texttt{"Please respond in <TARGET\_LANG>"} to the prompt. For training data of the base model, we concatenate the inputs and outputs and treat the result as the output sequence without applying the Q/A template.

The model is trained for one epoch using the LoRA fine-tuning technique~\cite{hu2022lora} with the following settings: \texttt{warmup\_ratio = 0.05}, \texttt{batch\_size = 1}, \texttt{gradient\_accumulation\_steps = 16}, \texttt{learning\_rate = 1e-4}, \texttt{lora\_rank = 8}, and \texttt{lora\_alpha = 32}. We employ the \texttt{MS-SWIFT} framework~\cite{zhao2024swiftascalablelightweightinfrastructure} for model training, with all other hyperparameters kept at their default values.

\subsection{Combination of ICL and \methodname{}}
We apply \methodname{} only to the current input prompt using the scaling factor specified in Appendix~\ref{app:itlc_exp_setting}, but not to the few-shot examples, since the languages used in the few-shot examples differ from that of the current task (see Appendix~\ref{app:icl_exp_setting}).

\subsection{Combination of PEFT and \methodname{}}
The LDA is trained based on hidden states extracted from the pre-trained model. We do not retrain the LDA using new hidden states after PEFT. For other details, please refer to Appendix~\ref{app:peft_exp_setting} and Appendix~\ref{app:itlc_exp_setting}

\subsection{INCLINE}
We extract sentence representations from the FLORES-200 dataset~\citep{nllbteam2022languageleftbehindscaling} and use the same scaling factor $\alpha$ as defined in Appendix~\ref{app:itlc_exp_setting} during inference.

\subsection{ReCoVeR}
We extract sentence representations from the FLORES-200 dataset~\citep{nllbteam2022languageleftbehindscaling} and apply a scaling factor of $\alpha = 0.2$ for Llama-3.1-8B and its instruct variant, and $\alpha = 0.3$ for Qwen2.5-0.5B and its instruct variant.

\section{Language Confusion Result}
\label{app:lang-confusion-result}

\subsection{Ablation Study of Scaling for Different Language Vector Injection Strategies} 
\label{app:ablation_scaling}
\begin{table}[!t]
\centering
\resizebox{\linewidth}{!}{
\begin{tabular}{lrrrrrr}
\toprule
Scaling & \multicolumn{3}{c}{Monolingual} & \multicolumn{3}{c}{Cross-lingual} \\
\cmidrule(lr){2-4} \cmidrule(lr){5-7}
& LCPR & LPR & WPR & LCPR & LPR & WPR \\
\midrule
prompt-0.1                     & 64.86 & 81.01 & 65.67 & 33.97 & 23.75 & 74.74 \\
prompt-0.2                     & \textbf{66.39} & 82.14 & \textbf{66.75} & 38.88 & 28.91 & \textbf{75.37} \\
prompt-0.3                     & 65.59 & \textbf{82.86} & 65.78 & 46.03 & 37.86 & 72.56 \\
prompt-0.4                     & 65.45 & 82.79 & 65.53 & 57.20 & 51.97 & 72.27 \\
prompt-0.5                     & 65.87 & 82.73 & 62.50 & 62.93 & 61.63 & 73.43 \\
prompt-0.6                     & 64.92 & 82.64 & 65.24 & 63.91 & 63.83 & 73.20 \\
prompt-0.7                     & 64.78 & 81.03 & 65.52 & 64.63 & 66.09 & 71.74 \\
prompt-0.8                     & 63.69 & 80.40 & 65.28 & \textbf{65.71} & \textbf{66.41} & 74.24 \\
prompt-0.9                     & 61.25 & 75.81 & 64.15 & 64.59 & 64.79 & 73.30 \\
prompt-1.0                     & 60.39 & 74.98 & 63.87 & 62.97 & 63.35 & 72.79 \\
\bottomrule
\end{tabular}
}

\caption{Performance (LCPR / LPR / WPR) of Qwen2.5-0.5B on LCB under the prompt-only setting with base shift vector, evaluated across different language vector scaling factors, $\alpha$.}
\label{tab:ablation-prompt-only}
\end{table}

\begin{table}[!t]
\centering
\resizebox{\linewidth}{!}{
\begin{tabular}{lrrrrrr}
\toprule
Scaling & \multicolumn{3}{c}{Monolingual} & \multicolumn{3}{c}{Cross-lingual} \\
\cmidrule(lr){2-4} \cmidrule(lr){5-7}
& LCPR & LPR & WPR & LCPR & LPR & WPR \\
\midrule
gen-0.1                     & 64.75 & 83.99 & 63.85 & 35.07 & 24.79 & 74.92 \\
gen-0.2                     & \textbf{65.35} & 85.09 & \textbf{65.01} & 39.93 & 28.96 & \textbf{75.92} \\
gen-0.3                     & 62.61 & 86.55 & 59.29 & 48.08 & 38.97 & 71.16 \\
gen-0.4                     & 59.61 & 86.23 & 54.95 & 57.49 & 57.82 & 64.37 \\
gen-0.5                     & 59.61 & 86.85 & 54.76 & 67.00 & 74.04 & 66.07 \\
gen-0.6                     & 60.05 & \textbf{87.49} & 58.14 & \textbf{71.35} & 80.46 & 67.67 \\
gen-0.7                     & 58.01 & 87.41 & 55.72 & 69.39 & \textbf{80.73} & 66.57 \\
gen-0.8                     & 52.45 & 82.78 & 52.35 & 65.84 & 75.74 & 65.93 \\
gen-0.9                     & 47.07 & 75.83 & 50.58 & 58.61 & 68.51 & 63.73 \\
gen-1.0                     & 40.44 & 71.15 & 54.91 & 51.25 & 61.85 & 61.83 \\
\bottomrule
\end{tabular}
}
\caption{Performance (LCPR / LPR / WPR) of Qwen2.5-0.5B on LCB under the generated-only setting with base shift vector, evaluated across different language vector scaling factors, $\alpha$.}
\label{tab:ablation-gen-only}
\end{table}

\begin{table}[!t]
\centering
\resizebox{\linewidth}{!}{
\begin{tabular}{lrrrrrr}
\toprule
Scaling & \multicolumn{3}{c}{Monolingual} & \multicolumn{3}{c}{Cross-lingual} \\
\cmidrule(lr){2-4} \cmidrule(lr){5-7}
& LCPR & LPR & WPR & LCPR & LPR & WPR \\
\midrule
prompt-and-gen-0.1                     & \textbf{64.21} & 84.27 & \textbf{63.77} & 39.48 & 28.69 & 75.74 \\
prompt-and-gen-0.2                     & 63.25 & 86.34 & 61.76 & 50.04 & 41.18 & 75.07 \\
prompt-and-gen-0.3                     & 62.94 & \textbf{88.24} & 60.85 & 64.22 & 64.18 & 72.53 \\
prompt-and-gen-0.4                     & 60.79 & 88.06 & 59.09 & 75.88 & 80.58 & 75.78 \\
prompt-and-gen-0.5                     & 59.98 & 87.11 & 59.41 & \textbf{78.93} & \textbf{85.08} & \textbf{77.15} \\
prompt-and-gen-0.6                     & 57.01 & 86.37 & 55.90 & 77.21 & 84.13 & 74.90 \\
prompt-and-gen-0.7                     & 53.56 & 82.91 & 53.63 & 72.57 & 81.98 & 71.51 \\
prompt-and-gen-0.8                     & 49.00 & 77.27 & 51.33 & 68.22 & 76.80 & 70.08 \\
prompt-and-gen-0.9                     & 40.41 & 70.51 & 48.16 & 60.97 & 69.07 & 66.44 \\
prompt-and-gen-1.0                     & 36.60 & 70.01 & 51.30 & 52.51 & 61.07 & 63.82 \\
\bottomrule
\end{tabular}
}
\caption{Performance (LCPR / LPR / WPR) of Qwen2.5-0.5B on LCB under the prompt-and-generated setting with base shift vector, evaluated across different language vector scaling factors, $\alpha$.}
\label{tab:ablation-prompt-and-gen}
\end{table}
 
As shown in Table~\ref{tab:ablation-prompt-only}, Table~\ref{tab:ablation-gen-only} and Table~\ref{tab:ablation-prompt-and-gen} Our analysis reveals distinct optimal scaling factors for cross-lingual LCPR across injection strategies: prompt-only achieves peak performance at scaling 0.8 (65.71), gen-only at 0.6 (71.35), and prompt-and-gen at 0.5 (78.93). Notably, prompt-and-gen outperforms other strategies, suggesting combined injection better preserves cross-lingual alignment. The scaling factor for the Qwen2.5-0.5B model family is adopted from our ablation study. However, due to computational constraints, a similar study was not feasible for the Qwen2.5-7B and Llama3.1-8B families. For these models, we instead conducted a limited manual evaluation, we randomly generated outputs for a range of scaling factors across different target languages and selected the best-performing value based on human assessment.

\subsection{Impact of In-context learning (ICL) on Monolingual and Cross-lingual Performance}

As shown in Table~\ref{tab:language-confusion-result-base}, Table~\ref{tab:language-confusion-result-instruct}, Table~\ref{tab:language-confusion-result-7b}, Table~\ref{tab:language-confusion-result-7b-instruct}, Table~\ref{tab:language-confusion-result-8b} and Table~\ref{tab:language-confusion-result-8b-instruct}, in the monolingual setting, the impact of few-shot prompting varies inconsistently across models. Qwen2.5-0.5B and Qwen2.5-0.5B-Instruct exhibit decreased LPR, while Qwen2.5-7B and Llama-3.1-8B show increased LPR. For instruction-tuned models, both Qwen2.5-7B-Instruct and Llama-3.1-8B-Instruct demonstrate reduced LPR. This unstable and unpredictable behavior may stem from the design of monolingual few-shot prompts, which introduce conflicting linguistic signals that models with limited capacity struggle to resolve effectively~\footnote{Please refer to Appendix~\ref{app:icl_exp_setting}.}.

In the cross-lingual setting, few-shot prompting consistently improves performance across all base models (Qwen2.5-0.5B, Qwen2.5-7B, and Llama-3.1-8B). This improvement can be attributed to the few-shot examples, which utilize English inputs paired with explicit target-language directives, thereby reinforcing the desired input-output alignment. These results indicate that English-centric prompting effectively stimulates cross-lingual adaptation in base models. However, the effect differs for instruction-tuned models: while smaller models like Qwen2.5-0.5B-Instruct benefit from few-shot examples, larger models (Qwen2.5-7B-Instruct and Llama-3.1-8B-Instruct) show minimal gains. This stability suggests that instruction-tuning pre-aligns their multilingual capabilities, rendering additional in-context examples largely redundant.

The divergent impact of ICL across models indicates that the effectiveness of few-shot prompting might contingent upon the model's instruction-following aptitude, contextual understanding, pre-existing upper-bound capability, and the depth of alignment achieved during its instruction-tuning process~\footnote{All discussed results are based on experiments that apply the official chat/QA templates during inference.}.

\subsection{Chat/QA Template Efficacy Across Settings}  
The findings are consistent with those observed in the in-context learning (ICL) setting for LPR performance, with one key exception: applying the chat template to instruction-tuned models consistently yields better performance, as shown in Table~\ref{tab:language-confusion-result-instruct}.

\subsection{Effect of Source Language Shift Vector}
\label{app:source_lang_substract}
\begin{figure}[!t]
\centering
\includegraphics[width=\linewidth]{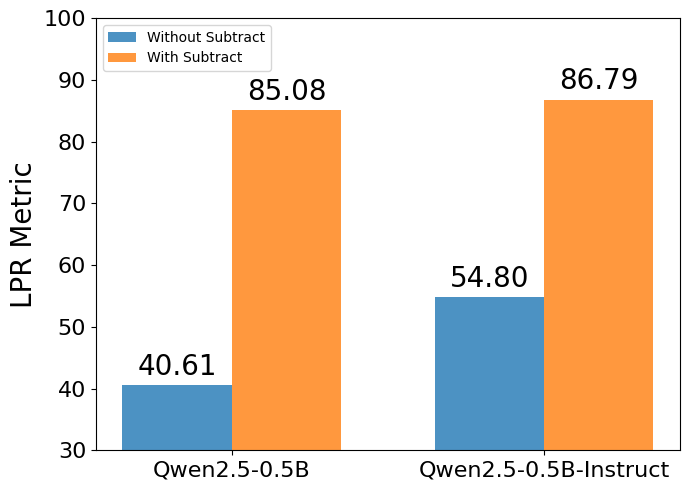}
\caption{Cross-lingual LPR performance on LCB with and without subtracting the source language shift vector across Qwen2.5-0.5B and Qwen2.5-0.5B-Instruct, using prompt-and-gen injection strategy with $\alpha=0.5$.}
\label{fig:substract-source-shift-vector}
\end{figure}

As shown in Figure~\ref{fig:substract-source-shift-vector}, subtracting the source language shift vector reduces the model's bias toward the source language (English) and guides the model to generate content in the target language more effectively, compared to directly adding the target language shift vector.

\begin{table}[!t]
\centering
\resizebox{\linewidth}{!}{
\begin{tabular}{lrrrrrr}
\toprule
Method & \multicolumn{3}{c}{Monolingual} & \multicolumn{3}{c}{Cross-lingual} \\
\cmidrule(lr){2-4} \cmidrule(lr){5-7}
& LCPR & LPR & WPR & LCPR & LPR & WPR \\
\midrule
Qwen2.5-0.5B                  & 65.27 & 81.58 & 65.15 & 29.41 & 19.75 & 73.45 \\
+ Q/A template (0-shot)       & 59.26 & 59.91 & 73.35 & 44.68 & 35.36 & 75.94 \\
\quad + PEFT                  & \textbf{75.96} & 82.91 & \textbf{78.30} & 76.15 & 77.55 & 80.56 \\
+ 1-shot                      & 56.12 & 55.38 & 73.70 & 47.42 & 37.95 & 75.42 \\
+ 2-shot                      & 51.59 & 49.70 & 70.98 & 49.36 & 41.64 & 75.03 \\
+ 3-shot                      & 52.52 & 51.51 & 72.07 & 53.16 & 46.65 & 77.07 \\
+ 4-shot                      & 54.16 & 52.95 & 74.15 & 55.03 & 48.23 & 77.60 \\
+ 5-shot                      & 54.47 & 53.62 & 70.40 & 56.78 & 50.63 & 76.16 \\
\midrule
+ \methodname{} (apply base shift vector) &               &       &       &       &       &       \\
\quad + prompt-only ($\alpha = 0.8$)                & 63.69 & 80.40 & 65.28 & 65.71 & 66.41 & 74.24 \\
\quad + gen-only ($\alpha = 0.6$)            & 60.05 & \textbf{87.49} & 58.14 & 71.35 & 80.46 & 67.67 \\
\quad + prompt-and-gen ($\alpha = 0.5$)            & 59.98 & 87.11 & 59.41 & {78.93} & 85.08 & 77.15 \\
\qquad + Q/A template & 62.50 & 81.21 & 64.60 & \textbf{81.30} & 85.61 & 80.84 \\
\qquad\quad + PEFT & 73.68 & 86.17 & 73.26 & 87.66 & \textbf{90.51} & \textbf{86.15} \\
\qquad + 5-shot & 57.65 & 74.38 & 61.13 & 81.51 & 87.58 & 79.01 \\
\midrule
+ \methodname{} (apply instruct shift vector) &               &       &       &       &       &       \\
\quad + prompt-only ($\alpha = 0.8$)                & 63.11 & 79.95 & 64.18 & 63.08 & 63.77 & 73.04 \\
\quad + gen-only ($\alpha = 0.6$)            & 55.89 & 86.38 & 55.32 & 68.70 & 78.99 & 65.36 \\
\quad + prompt-and-gen ($\alpha = 0.5$)            & 58.48 & 87.24 & 57.21 & 76.06 & 82.31 & 75.74 \\
\bottomrule
\end{tabular}
}
\caption{Performance (LCPR / LPR / WPR) of Qwen2.5-0.5B on LCB under monolingual and cross-lingual settings.}
\label{tab:language-confusion-result-base}
\end{table}

\begin{table}[!t]
\centering
\resizebox{\linewidth}{!}{
\begin{tabular}{lrrrrrr}
\toprule
Method & \multicolumn{3}{c}{Monolingual} & \multicolumn{3}{c}{Cross-lingual} \\
\cmidrule(lr){2-4} \cmidrule(lr){5-7}
& LCPR & LPR & WPR & LCPR & LPR & WPR \\
\midrule
Qwen2.5-0.5B-Instruct         & 74.79 & 82.61 & 77.94 & 38.75 & 27.22 & 78.40 \\
+ Chat template (0-shot)       & 74.52 & 83.66 & 77.12 & 63.00 & 57.69 & 79.50 \\
\quad + PEFT                  & \textbf{80.13} & 89.85 & 77.77 & 79.46 & 84.34 & 80.01 \\
+ 1-shot                      & 72.94 & 78.83 & 77.79 & 66.82 & 61.42 & 82.12 \\
+ 2-shot                      & 73.95 & 78.41 & 79.43 & 68.19 & 64.21 & 80.99 \\
+ 3-shot                      & 74.61 & 78.88 & 76.99 & 69.43 & 65.94 & 81.42 \\
+ 4-shot                      & 75.82 & 80.89 & \textbf{80.07} & 69.56 & 67.28 & 79.62 \\
+ 5-shot                      & 75.44 & 80.30 & 79.36 & 71.43 & 69.70 & 79.74 \\
\midrule
+ \methodname{} (apply base shift vector) &               &       &       &       &       &       \\
\quad + prompt-only ($\alpha = 0.8$)                & 67.33 & 74.82 & 76.35 & 76.05 & 77.68 & 81.11 \\
\quad + gen-only ($\alpha = 0.6$)            & 67.00 & 84.07 & 65.83 & 75.56 & 82.42 & 74.51 \\
\quad + prompt-and-gen ($\alpha = 0.5$)            & 67.73 & 81.70 & 68.96 & 81.51 & 85.32 & 80.55 \\
\midrule
+ \methodname{} (apply instruct shift vector) &               &       &       &       &       &       \\
\quad + prompt-only ($\alpha = 0.8$)                & 66.78 & 74.96 & 73.08 & 73.26 & 76.37 & 79.20 \\
\quad + gen-only ($\alpha = 0.6$)            & 67.42 & 83.64 & 65.46 & 73.95 & 84.06 & 71.40 \\
\quad + prompt-and-gen ($\alpha = 0.5$)            & 68.20 & 82.20 & 68.05 & 80.96 & 86.79 & 78.84 \\
\qquad + 5-shot                      & 68.93 & 86.28 & 66.47 & 83.98 & 88.07 & 82.00 \\
\qquad + PEFT                  & 68.16 & \textbf{90.51} & 62.58 & \textbf{85.38} & \textbf{89.85} & \textbf{82.83} \\
\bottomrule
\end{tabular}
}
\caption{Performance (LCPR / LPR / WPR) of Qwen2.5-0.5B-Instruct on LCB under monolingual and cross-lingual settings.}
\label{tab:language-confusion-result-instruct}
\end{table}

\begin{table}[!t]
\centering
\resizebox{\linewidth}{!}{
\begin{tabular}{lrrrrrr}
\toprule
Method & \multicolumn{3}{c}{Monolingual} & \multicolumn{3}{c}{Cross-lingual} \\
\cmidrule(lr){2-4} \cmidrule(lr){5-7}
& LCPR & LPR & WPR & LCPR & LPR & WPR \\
\midrule
Qwen2.5-7B         & 68.15 & 77.71 & 71.40 & 41.03 & 29.72 & 75.33 \\
+ Q/A template (0-shot)       & 53.97 & 55.24 & 73.84 & 65.68 & 60.61 & 76.88\\
\quad + PEFT & \textbf{73.46} & 83.80 & 72.80 & 78.93 & 82.66 & \textbf{79.51}\\
+ 5-shot           & 63.23 & 62.78 & \textbf{75.77} & 72.15 & 69.37 & 79.45 \\
\midrule
+ \methodname{} (apply base shift vector) &               &       &       &       &       &       \\
\quad + prompt-and-gen ($\alpha = 1.3$)            & 67.05 & 80.07 & 67.33 & 61.70 & 59.84 & 70.84 \\
\qquad + Q/A template  & 58.10 & 63.40 & 72.36 & 70.71 & 74.40 & 72.72 \\
\qquad\quad + PEFT & 73.12 & \textbf{85.60} & 72.40 & 78.25 & 83.92 & 78.39 \\
\quad + 5-shot & 65.24 & 69.55 & 73.42 & \textbf{79.60} & \textbf{84.90} & 77.13 \\
\bottomrule
\end{tabular}
}
\caption{Performance (LCPR / LPR / WPR) of Qwen2.5-7B on LCB under monolingual and cross-lingual settings.}
\label{tab:language-confusion-result-7b}
\end{table}

\begin{table}[!t]
\centering
\resizebox{\linewidth}{!}{
\begin{tabular}{lrrrrrr}
\toprule
Method & \multicolumn{3}{c}{Monolingual} & \multicolumn{3}{c}{Cross-lingual} \\
\cmidrule(lr){2-4} \cmidrule(lr){5-7}
& LCPR & LPR & WPR & LCPR & LPR & WPR \\
\midrule
Qwen2.5-7B-Instruct (with chat template)         & 60.83 & 78.89 & 58.78 & 66.16 & 78.81 & 62.37 \\
+ 5-shot                      & 54.46 & 74.13 & 53.93 & 65.79 & 78.51 & 61.44 \\
+ PEFT                      & 75.03 & 88.28 & \textbf{73.19} & \textbf{78.32} & 83.56 & \textbf{77.93}\\
\midrule
+ \methodname{} (apply base shift vector) &               &       &       &       &       &       \\
\quad + prompt-and-gen ($\alpha = 1.3$)              & 62.44 & 85.89 & 56.76 & 66.91 & 83.45 & 60.34 \\
\midrule
+ \methodname{} (apply instruct shift vector) &               &       &       &       &       &       \\
\quad + prompt-and-gen ($\alpha = 1.3$)              & 61.35 & 84.89 & 56.97 & 66.89 & \textbf{84.73} & 60.02 \\
\qquad + 5-shot & 57.75 & 81.01 & 53.73 & 66.26 & 84.04 & 58.97 \\
\qquad + PEFT & \textbf{75.62} & \textbf{90.12} & 72.33 & 77.50 & 84.10 & 76.70\\
\bottomrule
\end{tabular}
}
\caption{Performance (LCPR / LPR / WPR) of Qwen2.5-7B-Instruct on LCB under monolingual and cross-lingual settings.}
\label{tab:language-confusion-result-7b-instruct}
\end{table}

\begin{table}[!t]
\centering
\resizebox{\linewidth}{!}{
\begin{tabular}{lrrrrrr}
\toprule
Method & \multicolumn{3}{c}{Monolingual} & \multicolumn{3}{c}{Cross-lingual} \\
\cmidrule(lr){2-4} \cmidrule(lr){5-7}
& LCPR & LPR & WPR & LCPR & LPR & WPR \\
\midrule
Llama-3.1-8B         & 43.52 & 44.07 & 59.66 & 1.46 & 0.74 & \textbf{88.10} \\
+ Q/A template (0-shot)       & 63.68 & 56.98 & \textbf{82.26} & 39.01 & 26.13 & 87.27\\
\quad + PEFT & \textbf{79.16} & 93.01 & 72.80 & 82.04 & 89.73 & 77.83 \\
+ 5-shot           & 72.24 & 69.86 & 79.13 & 70.67 & 62.38 & 83.91 \\
\midrule
+ \methodname{} (apply base shift vector) &               &       &       &       &       &       \\
\quad + prompt-and-gen ($\alpha = 0.15$)            & 50.97 & 60.77 & 57.07 & 60.69 & 69.69 & 57.74 \\
\qquad + Q/A template & 73.13 & 75.77 & 77.28 & 81.29 & 81.68 & 82.78 \\
\qquad\quad + PEFT & 78.50 & \textbf{96.03} & 72.08 & \textbf{83.74} & \textbf{88.98} & 81.21 \\
\quad + 5-shot & 76.43 & 82.18 & 76.47 & 83.14 & 88.15 & 80.47 \\
\bottomrule
\end{tabular}
}
\caption{Performance (LCPR / LPR / WPR) of Llama-3.1-8B on LCB under monolingual and cross-lingual settings.}
\label{tab:language-confusion-result-8b}
\end{table}

\begin{table}[!t]
\centering
\resizebox{\linewidth}{!}{
\begin{tabular}{lrrrrrr}
\toprule
Method & \multicolumn{3}{c}{Monolingual} & \multicolumn{3}{c}{Cross-lingual} \\
\cmidrule(lr){2-4} \cmidrule(lr){5-7}
& LCPR & LPR & WPR & LCPR & LPR & WPR \\
\midrule
Llama-3.1-8B-Instruct (with chat template)         & 83.05 & 94.63 & 76.11 & 79.34 & 83.25 & 77.01 \\
+ 5-shot           & 82.27 & 88.57 & 79.88 & 84.32 & 86.68 & 82.77 \\
+ PEFT             & 79.00 & 96.66 & 71.00 & 81.26 & 91.13 & 75.29 \\
\midrule
+ \methodname{} (apply base shift vector) &               &       &       &       &       &       \\
\quad + prompt-and-gen ($\alpha = 0.10$)            & 82.50 & 95.68 & 75.68 & 83.48 & 88.52 & 80.37 \\
\midrule
+ \methodname{} (apply instruct shift vector) &               &       &       &       &       &       \\
\quad + prompt-and-gen ($\alpha = 0.10$)            & 81.76 & 96.41 & 74.51 & 82.91 & 89.06 & 78.99 \\
\qquad + 5-shot & \textbf{85.25} & 93.21 & \textbf{79.82} & \textbf{86.60} & 90.34 & \textbf{83.95} \\
\qquad + PEFT & 79.04 & \textbf{97.19} & 71.36 & 83.44 & \textbf{93.60} & 77.05 \\
\bottomrule
\end{tabular}
}
\caption{Performance (LCPR / LPR / WPR) of Llama-3.1-8B-Instruct on LCB under monolingual and cross-lingual settings.}
\label{tab:language-confusion-result-8b-instruct}
\end{table}

\section{Experiment setting for semantic retention and human evaluation}
\label{app:exp_set_lang_control}
\subsection{Generation Hyperparameter}
\label{app:generation-hyperparmater-setting}
The generation process for the language control and language confusion results uses specific hyperparameter to balance creativity and control. We set \texttt{max\_new\_tokens=50}, and set \texttt{top\_k} to 50. We apply nucleus sampling with \texttt{top\_p=0.9}, and use a moderate temperature of 0.7 to encourage focused yet varied outputs. To reduce repetitive phrases, we apply a \texttt{repetition\_penalty} of 1.5. We keep all other hyperparameters at their model-specific default values and use each instruct model's native chat template.

\subsection{Monolingual \& Crosslingual Prompting}
\label{app:monolingual-crosslingual-propmting}
Our experiments on the baseline (monolingual) and \methodname{} (cross-lingual) settings use slightly different prompt strategies. Specifically, for the baseline, we aim to measure the upper bound of performance within a particular language, whereas \methodname{} involves different input and target languages.

To ensure fairness and consistency in model output generation, we designed distinct input prompts for the base model, Qwen2.5, and Llama-3.1. In the base version, to control the contextual generation in cross-lingual settings, we prepend an early portion of the target language output—approximately 30\% of the sentence length—as a guidance signal for the model to continue generating coherent text. This approach helps ensure that the language vector receives sufficient signal to produce linguistically and semantically coherent outputs.

Additionally, for non-Latin scripts such as Japanese and Chinese, we adopt a different segmentation strategy. Instead of splitting based on newlines, as in Latin-script languages, we apply language-specific tokenizers such as PyThaiNLP \citep{phatthiyaphaibun-etal-2023-pythainlp}, Nagisa\footnote{\url{https://github.com/taishi-i/nagisa}}, and Jieba\footnote{\url{https://github.com/fxsjy/jieba}}. The proportional segment length is then determined based on the number of tokens or phrases produced by these tokenizers.

\section{Additional Examples of Cross-lingual Generation}
\label{app:example-generation-in-details}
Figure \ref{fig:example_xx2en} and Figure \ref{fig:example_en2xx} present several examples of generated outputs across multiple source languages targeting English. Overall, our \methodname{} method successfully shifts to the desired target language and demonstrates effective cross-lingual generation. 
\section{Annotation Guidelines}
\label{app:annotation-guideline}

\subsection{Context of the Annotation Task}

The annotation task involves evaluating the quality of cross-lingual language generation, where a model generates responses in a target language based on input prompts in a source language. The goal is to assess how well the model performs in terms of naturalness, relevance, and answer correctness. This evaluation is crucial for understanding the model's capabilities and identifying areas for improvement.

\subsection{Detailed Scoring Guidelines}

\subsubsection{Naturalness (1-5):}

\begin{itemize}
    \item \textbf{1:} The response sounds very unnatural, robotic, or translated. It lacks fluency and typical language patterns of the target language, making it sound artificial and unnatural.
    \item \textbf{2:} The response is somewhat unnatural, with noticeable awkwardness or unnatural word choices. It may sound stilted or forced.
    \item \textbf{3:} The response is moderately natural, with some minor awkwardness but generally understandable. It flows reasonably well but has room for improvement.
    \item \textbf{4:} The response is mostly natural, with only slight deviations from typical language use. It sounds almost native-like but may have minor imperfections.
    \item \textbf{5:} The response is completely natural, indistinguishable from text written by a native speaker. It flows smoothly and uses language patterns typical of the target language.
\end{itemize}

\subsubsection{Relevance (1-5):}

\begin{itemize}
    \item \textbf{1:} The response is completely irrelevant to the input prompt. It fails to address the topic or question posed.
    \item \textbf{2:} The response is somewhat relevant but misses key points or goes off-topic. It may touch on related ideas but does not fully address the prompt.
    \item \textbf{3:} The response is moderately relevant, addressing some aspects of the prompt but lacking completeness. It covers some key points but omits important details.
    \item \textbf{4:} The response is highly relevant, addressing most key points of the prompt. It provides a comprehensive answer but may miss minor details.
    \item \textbf{5:} The response is completely relevant, fully addressing all aspects of the prompt. It covers all key points and provides a thorough answer.
\end{itemize}

\subsubsection{Correctness (1-5):}

\begin{itemize}
    \item \textbf{1:} The response contains major factual errors or inaccuracies. It provides incorrect information or contradicts known facts.
    \item \textbf{2:} The response contains some factual errors or inaccuracies. It may be partially correct but includes misleading or incorrect details.
    \item \textbf{3:} The response is mostly correct but may have minor inaccuracies or omissions. It is generally accurate but requires minor corrections.
    \item \textbf{4:} The response is highly accurate, with only minor details potentially incorrect. It is reliable and trustworthy but may have small errors.
    \item \textbf{5:} The response is completely accurate and factually correct. It provides precise and reliable information without any errors.
\end{itemize}

\subsection{Additional Notes}

\begin{itemize}
    \item \textbf{Contextual Understanding:} Annotators should consider the context of the input prompt and the intended audience when evaluating naturalness and relevance. A response that is natural and relevant in one context may not be in another.
    \item \textbf{Consistency:} Annotators should strive for consistency in their annotations across different examples. This helps ensure that the evaluation is fair and reliable.
    \item \textbf{Examples:} Providing clear examples of each rating level for each category can help annotators understand the expected standards and make consistent judgments.
    \item \textbf{Feedback:} Encourage annotators to provide feedback on ambiguous cases or areas where the guidelines could be improved. This can help refine the annotation process and improve the quality of the evaluations.
\end{itemize}

\section{Authors’ Contributions}
\label{app:author_contribution}
Joanito Agili Lopo is primarily responsible for developing the methodology as well as conducting experiments related to semantic retention and human evaluation. Muhammad Ravi Shulthan Habibi is responsible for the representation alignment and language identification (LID) experiments using linear probing. Tack Hwa Wong co-developed the methodology with Joanito Agili Lopo and is responsible for the LID experiments using k-nearest neighbours (KNN), as well as all experiments related to language confusion and language confusion benchmark (LCB). Samuel Cahyawijaya provided the research topic, overall direction, ideas, and guidance throughout the entire work. All co–first authors contributed to the writing of the first draft, and all authors participated in the review and editing process.

\begin{table}[!htbp]
\centering
\footnotesize
\setlength{\tabcolsep}{4pt}
\definecolor{lightgray}{gray}{0.9}
\newcolumntype{g}{>{\columncolor{lightgray}}c} 
\resizebox{\linewidth}{!}{
\begin{tabular}{l g c c c c c c}
\toprule
& \multicolumn{1}{c}{\cellcolor{white}} & \multicolumn{5}{c}{\textbf{Cross-lingual}} \\
\cmidrule(lr){2-8}
\textbf{Model} & \textbf{AVG} & \textbf{AR} & \textbf{ES} & \textbf{HI} & \textbf{ID} & \textbf{RU} & \textbf{ZH} \\
\midrule
Qwen2.5-0.5B  & 34.97 & 31.72 & 48.12 & 3.03 & 42.44 & 48.77 & 35.74 \\
+ INCLINE              & 43.82 & 34.94 & 74.17 & 6.58 & 56.38 & 59.22 & 31.63 \\
+ ReCoVeR              & 88.43 & 99.66 & 97.02 & 64.67 & 84.88 & 98.99 & 85.38 \\
+ \methodname{} (ours) & 81.22 & 98.32 & 94.61 & 32.32 & 83.17 & 97.65 & 81.25 \\
\midrule
Llama-3.1-8B  & 25.05 & 10.60 & 37.63 & 25.71 & 38.13 & 17.61 & 20.59 \\
+ INCLINE              & 34.69 & 19.61 & 39.25 & 38.92 & 40.46 & 32.36 & 37.56 \\
+ ReCoVeR              & 88.79 & 100.00 & 84.30 & 93.44 & 70.97 & 98.69 & 85.37 \\
+ \methodname{} (ours) & 76.38 & 90.41 & 83.57 & 76.43 & 62.37 & 97.29 & 48.24 \\
\bottomrule
\end{tabular}
}
\caption{LPR metrics for the base model on LCB across baseline and state-of-the-art methods, with a detailed language-wise breakdown for cross-lingual settings. All results have been applied with the QA/Chat template during inference.}
\label{tab:lpr-intervention-per-lang-base}
\end{table}

\newpage
\begin{table}[!htbp]
\centering
\footnotesize
\setlength{\tabcolsep}{4pt}
\definecolor{lightgray}{gray}{0.9}
\newcolumntype{g}{>{\columncolor{lightgray}}c} 
\resizebox{\linewidth}{!}{
\begin{tabular}{l g c c c c c c}
\toprule
& \multicolumn{1}{c}{\cellcolor{white}} & \multicolumn{5}{c}{\textbf{Cross-lingual}} \\
\cmidrule(lr){2-8}
\textbf{Model} & \textbf{AVG} & \textbf{AR} & \textbf{ES} & \textbf{HI} & \textbf{ID} & \textbf{RU} & \textbf{ZH} \\
\midrule
Qwen2.5-0.5B-Instruct  & 52.28 & 65.41 & 72.65 & 3.02 & 54.35 & 77.12 & 41.14 \\
+ INCLINE              & 56.54 & 68.35 & 80.35 & 1.13 & 52.19 & 68.08 & 69.16 \\
+ ReCoVeR              & 84.21 & 100.00 & 97.66 & 60.36 & 58.86 & 99.31 & 89.04 \\
+ \methodname{} (ours) & 81.97 & 98.97 & 95.31 & 49.03 & 64.39 & 98.98 & 85.13 \\
\midrule
Llama-3.1-8B-Instruct  & 80.68 & 87.12 & 89.27 & 82.76 & 73.89 & 87.93 & 63.14 \\
+ INCLINE              & 80.63 & 86.80 & 89.60 & 81.10 & 70.21 & 86.58 & 69.51 \\
+ ReCoVeR              & 90.29 & 100.00 & 93.30 & 95.24 & 67.96 & 99.32 & 85.92 \\
+ \methodname{} (ours) & 85.65 & 95.60 & 92.96 & 93.97 & 72.55 & 95.98 & 62.84 \\
\bottomrule
\end{tabular}
}
\caption{LPR metrics for the instruct model on LCB across baseline and state-of-the-art methods, with a detailed language-wise breakdown for cross-lingual settings. All results have been applied with the QA/Chat template during inference.}
\label{tab:lpr-intervention-per-lang-instruct}
\end{table}

\begin{table*}[t]
  \centering
  \footnotesize
  \setlength{\tabcolsep}{4pt}
  \begin{adjustbox}{width=\textwidth,center}
    \definecolor{lightgray}{gray}{0.9}
    \newcolumntype{g}{>{\columncolor{lightgray}}c}
    \begin{tabular}{l g *{15}{c}}
      \toprule
      & \multicolumn{15}{c}{\textbf{Monolingual}} \\
      \cmidrule(lr){2-17}
      \textbf{Model} & {\textbf{AVG}} & {\textbf{AR}} & {\textbf{DE}} & {\textbf{EN}} & {\textbf{ES}} & {\textbf{FR}} & {\textbf{HI}} & {\textbf{ID}} & {\textbf{IT}} & {\textbf{JA}} & {\textbf{KO}} & {\textbf{PT}} & {\textbf{RU}} & {\textbf{TR}} & {\textbf{VI}} & {\textbf{ZH}} \\
      \midrule
      Qwen2.5-0.5B      & 59.91 & 45.84 & 78.79 & 97.00 & 75.20 & 64.67 & 0.00 & 57.00 & 76.00 & 32.00 & 54.55 & 64.00 & 64.29 & 30.00 & 81.82 & 77.50 \\
      + ICL (5-shot) & 53.62 & 56.12 & 84.00 & 96.44 & 64.86 & 54.53 & 4.17 & 64.00 & 65.66 & 19.19 & 40.40 & 45.00 & 73.74 & 25.00 & 68.37 & 42.81 \\
      \quad + \methodname{} (ours)  & 74.38 & 77.94 & 94.00 & 99.49 & 94.33 & 89.67 & 0.00 & 78.00 & 77.00 & 55.56 & 74.75 & 79.50 & 87.00 & 55.00 & 74.00 & 79.50 \\
      + PEFT           & 82.91 & 94.00 & 99.00 & 70.50 & 93.00 & 94.67 & 0.00 & 90.00 & 98.00 & 64.00 & 86.00 & 91.00 & 95.00 & 92.00 & 94.00 & 82.50 \\
      \quad + \methodname{} (ours)          & 86.17 & 99.33 & 100.00 & 77.00 & 99.33 & 99.33 & 8.25 & 94.00 & 100.00 & 57.00 & 81.82 & 98.00 & 98.00 & 99.00 & 90.00 & 91.50 \\
      + \methodname{} (ours) & 81.21 & 91.00 & 96.00 & 97.98 & 98.67 & 98.00 & 0.00 & 84.00 & 100.00 & 58.00 & 81.00 & 95.00 & 81.00 & 75.00 & 70.00 & 92.50 \\
      \midrule
      Qwen2.5-7B      & 55.24 & 29.43 & 73.00 & 98.48 & 70.04 & 66.17 & 1.01 & 63.00 & 78.00 & 39.00 & 22.68 & 65.00 & 36.08 & 26.80 & 83.00 & 76.85 \\
      + ICL (5-shot) & 62.78 & 43.26 & 79.00 & 96.39 & 71.84 & 74.02 & 15.96 & 73.74 & 82.00 & 59.00 & 44.79 & 50.36 & 65.66 & 56.25 & 82.00 & 47.50 \\
      \quad + \methodname{} (ours)  & 69.55 & 51.22 & 86.87 & 97.94 & 77.44 & 82.25 & 8.70 & 86.00 & 91.00 & 64.00 & 53.54 & 64.50 & 77.55 & 56.25 & 89.00 & 57.00 \\
      + PEFT           & 83.80 & 95.00 & 99.00 & 49.58 & 94.00 & 94.00 & 6.06 & 91.00 & 97.98 & 75.00 & 85.00 & 91.94 & 96.00 & 94.00 & 100.00 & 88.50 \\
      \quad + \methodname{} (ours)          & 85.60 & 98.67 & 99.00 & 52.97 & 97.67 & 95.67 & 8.00 & 95.00 & 94.00 & 76.00 & 89.00 & 95.00 & 97.00 & 97.00 & 100.00 & 89.00 \\
      + \methodname{} (ours) & 63.40 & 52.79 & 76.00 & 98.99 & 84.71 & 77.53 & 0.00 & 75.00 & 78.00 & 45.92 & 31.00 & 77.44 & 60.61 & 21.65 & 85.86 & 85.50 \\
      \midrule
      Llama-3.1-8B      & 56.98 & 39.57 & 55.00 & 95.38 & 69.56 & 59.43 & 30.21 & 57.58 & 55.56 & 25.51 & 43.30 & 71.29 & 67.37 & 81.82 & 61.00 & 42.19 \\
      + ICL (5-shot) & 69.86 & 67.53 & 75.00 & 95.47 & 69.33 & 63.67 & 63.64 & 73.00 & 73.00 & 67.00 & 49.00 & 67.82 & 70.00 & 70.00 & 76.00 & 67.50 \\
      \quad + \methodname{} (ours)  & 82.18 & 79.26 & 90.00 & 99.50 & 92.67 & 84.00 & 65.00 & 66.00 & 90.00 & 87.00 & 68.37 & 86.92 & 89.00 & 70.00 & 87.00 & 78.00 \\
      + PEFT           & 93.01 & 98.00 & 98.00 & 69.50 & 94.67 & 92.00 & 92.00 & 84.00 & 99.00 & 94.00 & 93.00 & 93.50 & 97.00 & 95.00 & 98.00 & 97.50 \\
      \quad + \methodname{} (ours)          & 96.03 & 100.00 & 97.00 & 91.50 & 97.67 & 97.33 & 96.00 & 91.00 & 99.00 & 95.00 & 97.00 & 92.50 & 99.00 & 99.00 & 98.00 & 90.50 \\
      + \methodname{} (ours) & 75.77 & 62.08 & 78.00 & 99.00 & 89.29 & 84.02 & 50.00 & 66.67 & 81.00 & 58.76 & 76.84 & 85.78 & 93.81 & 86.73 & 75.51 & 49.00 \\
      \midrule
      & \multicolumn{15}{c}{\textbf{Cross-lingual}} \\
      \cmidrule(lr){2-17}
      \textbf{Model} & {\textbf{AVG}} & {\textbf{AR}} & {\textbf{DE}} & {\textbf{EN}} & {\textbf{ES}} & {\textbf{FR}} & {\textbf{HI}} & {\textbf{ID}} & {\textbf{IT}} & {\textbf{JA}} & {\textbf{KO}} & {\textbf{PT}} & {\textbf{RU}} & {\textbf{TR}} & {\textbf{VI}} & {\textbf{ZH}} \\
      \midrule
      Qwen2.5-0.5B      & 35.36 & 31.72 & 43.27 & -- & 48.12 & 46.45 & 3.03 & 42.44 & 40.33 & 14.40 & 10.12 & 45.11 & 48.77 & 34.23 & 51.28 & 35.74 \\
      + ICL (5-shot) & 50.63 & 54.79 & 63.97 & -- & 54.62 & 63.02 & 12.07 & 61.97 & 63.05 & 24.74 & 29.57 & 55.90 & 67.84 & 61.61 & 69.21 & 26.38 \\
      \quad + \methodname{} (ours)  & 87.58 & 99.66 & 97.99 & -- & 96.62 & 97.33 & 39.07 & 85.52 & 95.26 & 72.91 & 88.95 & 90.95 & 98.99 & 91.96 & 92.63 & 78.24 \\
      + PEFT           & 77.55 & 89.25 & 90.26 & -- & 90.94 & 90.94 & 11.04 & 75.41 & 88.25 & 68.52 & 65.32 & 82.23 & 90.94 & 83.53 & 90.26 & 68.84 \\
      \quad + \methodname{} (ours)          & 90.51 & 100.00 & 99.67 & -- & 96.65 & 97.32 & 63.78 & 85.61 & 98.99 & 69.22 & 88.97 & 90.97 & 99.67 & 96.99 & 96.99 & 82.26 \\
      + \methodname{} (ours) & 85.61 & 98.32 & 96.97 & -- & 94.61 & 95.63 & 32.32 & 83.17 & 99.00 & 61.20 & 82.55 & 88.28 & 97.65 & 92.96 & 94.60 & 81.25 \\
      \midrule
      Qwen2.5-7B      & 60.61 & 62.24 & 67.82 & -- & 71.07 & 68.68 & 24.87 & 60.80 & 67.31 & 51.90 & 50.29 & 68.40 & 69.21 & 59.40 & 72.07 & 54.42 \\
      + ICL (5-shot) & 69.37 & 70.22 & 77.42 & -- & 75.04 & 75.20 & 36.45 & 70.53 & 81.43 & 59.16 & 59.26 & 70.02 & 84.24 & 77.05 & 79.20 & 55.94 \\
      \quad + \methodname{} (ours) & 84.90 & 88.57 & 95.50 & -- & 90.40 & 92.14 & 65.67 & 84.03 & 90.37 & 57.86 & 85.17 & 88.74 & 94.48 & 91.58 & 90.92 & 73.18 \\
      + PEFT           & 82.66 & 93.62 & 93.23 & -- & 89.27 & 89.93 & 24.20 & 83.16 & 86.25 & 76.87 & 80.56 & 86.84 & 95.29 & 91.57 & 90.93 & 75.53 \\
      \quad + \methodname{} (ours)          & 83.92 & 97.65 & 97.95 & -- & 96.99 & 95.31 & 30.78 & 87.60 & 93.97 & 35.09 & 74.47 & 92.59 & 97.65 & 96.99 & 96.98 & 80.89 \\
      + \methodname{} (ours) & 74.40 & 83.00 & 89.49 & -- & 89.51 & 87.43 & 27.12 & 76.65 & 87.42 & 32.80 & 58.81 & 87.35 & 91.49 & 82.97 & 85.93 & 61.61 \\
      \midrule
      Llama-3.1-8B      & 26.13 & 10.60 & 28.03 & -- & 37.63 & 36.09 & 25.71 & 38.13 & 37.14 & 18.88 & 16.49 & 31.77 & 17.61 & 20.14 & 27.05 & 20.59 \\
      + ICL (5-shot) & 62.38 & 65.02 & 60.66 & -- & 66.88 & 56.64 & 65.72 & 71.81 & 65.46 & 46.49 & 68.77 & 56.07 & 69.50 & 73.40 & 63.12 & 43.83 \\
      \quad + \methodname{} (ours)  & 88.15 & 85.24 & 96.97 & -- & 87.62 & 84.40 & 76.23 & 76.51 & 87.56 & 93.79 & 96.94 & 89.34 & 99.66 & 92.87 & 92.58 & 74.44 \\
      + PEFT           & 89.73 & 93.61 & 92.27 & -- & 91.28 & 93.64 & 93.62 & 76.16 & 89.60 & 85.57 & 85.50 & 89.22 & 94.24 & 92.28 & 94.30 & 84.90 \\
      \quad + \methodname{} (ours)          & 88.98 & 98.99 & 96.96 & -- & 86.21 & 75.21 & 98.65 & 67.22 & 89.96 & 88.95 & 95.61 & 84.58 & 99.33 & 95.31 & 92.96 & 75.86 \\
      + \methodname{} (ours) & 81.68 & 90.41 & 96.13 & -- & 83.57 & 71.68 & 76.43 & 62.37 & 89.12 & 75.72 & 89.11 & 82.56 & 97.29 & 87.75 & 93.09 & 48.24 \\
      \bottomrule
    \end{tabular}
  \end{adjustbox}
  \caption{LPR metrics for the base model on LCB, with a detailed language-wise breakdown for both monolingual and cross-lingual settings. All results have been applied with the QA/Chat template during inference.}
  \label{tab:lcpr-per-lang-base}
\end{table*}

\begin{table*}[t]
  \centering
  \footnotesize
  \setlength{\tabcolsep}{4pt}
  \begin{adjustbox}{width=\textwidth,center}
    \definecolor{lightgray}{gray}{0.9}
    \newcolumntype{g}{>{\columncolor{lightgray}}c}
    \begin{tabular}{l g *{15}{c}}
      \toprule
      & \multicolumn{15}{c}{\textbf{Monolingual}} \\
      \cmidrule(lr){2-17}
      \textbf{Model} & {\textbf{AVG}} & {\textbf{AR}} & {\textbf{DE}} & {\textbf{EN}} & {\textbf{ES}} & {\textbf{FR}} & {\textbf{HI}} & {\textbf{ID}} & {\textbf{IT}} & {\textbf{JA}} & {\textbf{KO}} & {\textbf{PT}} & {\textbf{RU}} & {\textbf{TR}} & {\textbf{VI}} & {\textbf{ZH}} \\
      \midrule
      Qwen2.5-0.5B-Instruct        & 83.66 & 96.33 & 94.00 & 99.50 & 89.67 & 95.33 & 0.00 & 70.00 & 94.00 & 82.00 & 83.51 & 87.00 & 95.00 & 89.00 & 87.63 & 92.00 \\
      + ICL (5-shot)   & 80.30 & 93.56 & 95.00 & 97.50 & 87.67 & 89.67 & 2.04 & 69.00 & 94.00 & 67.00 & 78.72 & 83.50 & 89.90 & 86.00 & 87.88 & 83.00 \\
      \quad + \methodname{} (ours)  & 86.28 & 98.33 & 98.00 & 98.50 & 97.67 & 96.67 & 13.00 & 82.00 & 98.00 & 80.00 & 77.00 & 94.00 & 89.00 & 96.00 & 95.00 & 81.00 \\
      + PEFT           & 89.85 & 99.00 & 99.00 & 96.50 & 95.67 & 97.67 & 14.43 & 87.00 & 100.00 & 83.00 & 93.94 & 95.50 & 100.00 & 95.00 & 99.00 & 92.00 \\
      \quad + \methodname{} (ours)          & 90.51 & 100.00 & 98.00 & 100.00 & 98.67 & 100.00 & 29.00 & 94.00 & 100.00 & 80.00 & 81.00 & 98.00 & 88.00 & 99.00 & 99.00 & 93.00 \\
      + \methodname{} (ours) & 82.20 & 100.00 & 99.00 & 100.00 & 98.67 & 98.33 & 7.00 & 74.00 & 100.00 & 80.00 & 72.00 & 95.50 & 39.00 & 95.00 & 82.00 & 92.50 \\
      \midrule
      Qwen2.5-7B-Instruct      & 78.89 & 81.03 & 96.00 & 95.49 & 87.17 & 87.97 & 31.58 & 72.00 & 91.00 & 55.00 & 61.54 & 84.50 & 81.32 & 88.89 & 87.88 & 82.00 \\
      + ICL (5-shot) & 74.13 & 70.08 & 91.92 & 90.91 & 77.12 & 83.72 & 38.46 & 64.65 & 84.85 & 50.00 & 68.66 & 79.72 & 71.28 & 77.66 & 86.87 & 76.00 \\
      \quad + \methodname{} (ours)  & 81.01 & 80.85 & 92.00 & 92.88 & 86.68 & 86.45 & 51.61 & 68.00 & 87.88 & 75.00 & 81.32 & 83.27 & 71.58 & 90.43 & 84.21 & 83.00 \\
      + PEFT           & 88.28 & 97.66 & 92.00 & 99.00 & 93.30 & 94.56 & 13.40 & 88.00 & 97.00 & 84.00 & 84.38 & 95.00 & 94.95 & 99.00 & 99.00 & 93.00 \\
      \quad + \methodname{} (ours)          & 90.12 & 99.33 & 98.00 & 98.49 & 96.99 & 96.00 & 20.20 & 89.00 & 96.00 & 80.00 & 91.75 & 96.50 & 97.00 & 97.00 & 99.00 & 96.50 \\
      + \methodname{} (ours) & 84.89 & 89.29 & 96.00 & 95.50 & 91.91 & 94.28 & 42.11 & 76.77 & 92.00 & 72.00 & 81.32 & 87.00 & 82.47 & 92.78 & 89.90 & 90.00 \\
      \midrule
      Llama-3.1-8B-Instruct      & 94.63 & 97.00 & 99.00 & 98.00 & 95.67 & 95.33 & 90.00 & 82.00 & 97.00 & 95.00 & 89.00 & 91.00 & 98.00 & 100.00 & 100.00 & 92.50 \\
      + ICL (5-shot) & 88.57 & 93.33 & 99.00 & 16.50 & 95.67 & 96.33 & 92.00 & 89.00 & 97.00 & 86.00 & 96.00 & 89.50 & 94.00 & 94.79 & 100.00 & 89.50 \\
      \quad + \methodname{} (ours)  & 93.21 & 97.00 & 98.00 & 46.74 & 96.00 & 98.00 & 99.00 & 90.00 & 99.00 & 95.00 & 98.00 & 95.00 & 99.00 & 92.86 & 100.00 & 94.50 \\
      + PEFT           & 96.66 & 98.67 & 97.00 & 97.50 & 95.33 & 98.00 & 96.00 & 95.00 & 99.00 & 91.00 & 91.00 & 97.00 & 95.96 & 100.00 & 100.00 & 98.50 \\
      \quad + \methodname{} (ours)          & 97.19 & 100.00 & 100.00 & 98.99 & 97.67 & 97.67 & 95.00 & 89.00 & 100.00 & 93.00 & 94.00 & 96.50 & 98.00 & 100.00 & 100.00 & 98.00 \\
      + \methodname{} (ours) & 96.41 & 99.33 & 99.00 & 99.00 & 96.33 & 98.00 & 94.00 & 88.00 & 99.00 & 92.00 & 97.00 & 94.50 & 100.00 & 98.00 & 100.00 & 92.00 \\
      \midrule
      & \multicolumn{15}{c}{\textbf{Cross-lingual}} \\
      \cmidrule(lr){2-17}
      \textbf{Model} & {\textbf{AVG}} & {\textbf{AR}} & {\textbf{DE}} & {\textbf{EN}} & {\textbf{ES}} & {\textbf{FR}} & {\textbf{HI}} & {\textbf{ID}} & {\textbf{IT}} & {\textbf{JA}} & {\textbf{KO}} & {\textbf{PT}} & {\textbf{RU}} & {\textbf{TR}} & {\textbf{VI}} & {\textbf{ZH}} \\
      \midrule
      Qwen2.5-0.5B-Instruct        & 57.69 & 65.41 & 72.12 & -- & 72.65 & 71.82 & 3.02 & 54.35 & 63.95 & 45.09 & 39.18 & 68.47 & 77.12 & 62.79 & 70.60 & 41.14 \\
      + ICL (5-shot)   & 69.70 & 81.82 & 83.57 & -- & 79.01 & 80.73 & 8.38 & 67.25 & 80.70 & 61.51 & 63.66 & 73.39 & 83.97 & 79.14 & 75.93 & 56.71 \\
      \quad + \methodname{} (ours)  & 88.07 & 100.00 & 97.98 & -- & 95.93 & 93.27 & 64.93 & 65.85 & 95.29 & 79.26 & 87.21 & 87.81 & 99.00 & 96.63 & 97.99 & 71.78 \\
      + PEFT           & 84.34 & 91.72 & 92.75 & -- & 93.50 & 93.16 & 14.45 & 85.75 & 94.12 & 85.07 & 77.16 & 90.56 & 95.55 & 90.59 & 96.50 & 79.85 \\
      \quad + \methodname{} (ours)          & 89.85 & 100.00 & 98.65 & -- & 95.95 & 93.95 & 61.88 & 76.10 & 96.94 & 77.24 & 86.95 & 92.65 & 98.98 & 96.29 & 98.30 & 84.00 \\
      + \methodname{} (ours) & 86.79 & 98.97 & 97.64 & -- & 95.31 & 92.27 & 49.03 & 64.39 & 97.31 & 73.24 & 79.27 & 91.98 & 98.98 & 93.25 & 98.32 & 85.13 \\
      \midrule
      Qwen2.5-7B-Instruct     & 78.81 & 81.96 & 88.38 & -- & 83.92 & 84.49 & 52.64 & 73.14 & 82.92 & 71.04 & 79.19 & 85.16 & 80.32 & 88.54 & 77.73 & 73.85 \\
      + ICL (5-shot) & 78.51 & 79.33 & 92.24 & -- & 84.59 & 86.68 & 58.27 & 66.12 & 87.17 & 67.86 & 78.04 & 80.94 & 85.12 & 84.47 & 73.78 & 74.52 \\
      \quad + \methodname{} (ours)  & 84.04 & 86.46 & 96.95 & -- & 87.60 & 91.83 & 62.21 & 70.94 & 91.90 & 69.84 & 87.71 & 87.16 & 90.19 & 92.39 & 85.86 & 75.53 \\
      + PEFT           & 83.56 & 94.54 & 91.04 & -- & 91.73 & 91.22 & 27.18 & 82.67 & 87.85 & 79.03 & 82.33 & 87.08 & 95.29 & 88.69 & 91.48 & 79.72 \\
      \quad + \methodname{} (ours)          & 84.10 & 98.65 & 98.26 & -- & 94.55 & 96.23 & 26.55 & 84.20 & 96.29 & 37.19 & 80.42 & 92.61 & 96.64 & 95.22 & 96.31 & 84.23 \\
      + \methodname{} (ours) & 84.73 & 86.71 & 95.21 & -- & 90.56 & 91.19 & 57.03 & 75.07 & 94.23 & 63.87 & 88.74 & 87.56 & 92.88 & 92.92 & 90.95 & 79.27 \\
      \midrule
      Llama-3.1-8B-Instruct      & 83.25 & 87.12 & 89.92 & -- & 89.27 & 85.58 & 82.76 & 73.89 & 89.25 & 71.51 & 80.10 & 82.57 & 87.93 & 90.52 & 91.94 & 63.14 \\
      + ICL (5-shot) & 86.68 & 86.24 & 91.60 & -- & 89.60 & 91.94 & 86.17 & 74.53 & 90.26 & 81.89 & 90.80 & 80.90 & 92.18 & 90.26 & 94.29 & 72.83 \\
      \quad + \methodname{} (ours)  & 90.34 & 92.62 & 97.32 & -- & 93.63 & 90.88 & 95.30 & 71.87 & 94.27 & 89.95 & 95.96 & 85.53 & 96.31 & 94.63 & 93.63 & 72.86 \\
      + PEFT           & 91.13 & 95.22 & 94.18 & -- & 95.30 & 94.96 & 92.22 & 79.09 & 94.20 & 87.18 & 89.06 & 86.86 & 93.82 & 91.28 & 93.57 & 88.84 \\
      \quad + \methodname{} (ours)          & 93.60 & 97.54 & 96.96 & -- & 94.64 & 94.59 & 96.93 & 80.14 & 93.91 & 93.96 & 94.54 & 91.29 & 96.94 & 96.26 & 96.27 & 86.46 \\
      + \methodname{} (ours) & 89.06 & 95.60 & 97.99 & -- & 92.96 & 93.64 & 93.97 & 72.55 & 92.62 & 83.60 & 91.98 & 83.94 & 95.98 & 93.95 & 95.30 & 62.84 \\
      \bottomrule
    \end{tabular}
  \end{adjustbox}
  \caption{LPR metrics for the instruct model on LCB, with a detailed language-wise breakdown for both monolingual and cross-lingual settings. All results have been applied with the QA/Chat template during inference.}
  \label{tab:lcpr-per-lang-instruct}
\end{table*}

\end{document}